\documentclass[preprint,12pt]{elsarticle}

\PassOptionsToPackage{square,numbers,sort&compress}{natbib}

\usepackage[T1]{fontenc}
\usepackage[utf8]{inputenc}
\usepackage{microtype}
\usepackage{amsmath,amssymb}

\usepackage{graphicx}
\usepackage{booktabs, array, multirow, tabularx, makecell, xcolor}

\usepackage{siunitx}
\usepackage[ruled,vlined,linesnumbered]{algorithm2e}
\usepackage{algorithmic}

\usepackage{tikz}
\usetikzlibrary{arrows.meta, matrix, plotmarks}
\usepackage{pgfplots}
\pgfplotsset{compat=1.18}

\usepackage{hyperref}
\usepackage[nameinlink,noabbrev]{cleveref}

\usepackage{float}
\usepackage{appendix}

\usepackage{tikz}
\usepackage{pgfplots}
\pgfplotsset{compat=1.18}
\usetikzlibrary{arrows.meta,plotmarks}

\usepackage{caption}
\usepackage{booktabs, array, tabularx}
\newcolumntype{Y}{>{\centering\arraybackslash}X}
\usepackage{graphicx}
\usepackage[caption=false]{subfig}  
\providecommand{\Description}[1]{} 

\usepackage{booktabs, array, tabularx}
\newcolumntype{Y}{>{\centering\arraybackslash}X}

\DeclareUnicodeCharacter{2013}{--}

\hypersetup{
  colorlinks=true,
  linkcolor=black, citecolor=black, urlcolor=black,
  pdftitle={A Reinforcement-Driven Multiple Instance Learning Method for Multi-Task Speaker Attribute Prediction},
  pdfauthor={Stijn Lakeman; Seyed Sahand Mohammadi Ziabari; Ali Mohammed Mansoor Alsahag}
}

\journal{<Annals of Data Science>} 

\begin{document}

\begin{frontmatter}

\title{Fairness Metric Design Exploration in Multi-Domain Moral Sentiment Classification using Transformer-Based Models}

\author[inst1]{Battemuulen Naranbat}
\ead{battemuulen.naranbat@student.uva.nl}

\author[inst1,inst2]{Seyed Sahand Mohammadi Ziabari\corref{cor1}}
\ead{sahand.ziabari@sunyempire.edu}

\author[inst3]{Yousuf Nasser Al Husaini}
\ead{yousufnaser@aou.edu.om}

\author[inst1]{Ali Mohammed Mansoor Alsahag}
\ead{a.m.m.alsahag@uva.nl}

\cortext[cor1]{Corresponding author.}

\affiliation[inst1]{organization={Informatics Institute, University of Amsterdam},
  addressline={Science Park}, postcode={1098 XH}, city={Amsterdam}, country={The Netherlands}}

\affiliation[inst2]{organization={Department of Computer Science and Technology, SUNY Empire State University},
  city={Saratoga Springs}, state={NY}, country={USA}}

\affiliation[inst3]{organization={Faculty of Computer Studies, Arab Open University, Oman},
  addressline={P.O. Box 1569}, postcode={130}, city={Muscat}, country={Sultanate of Oman}}

\begin{abstract}
Ensuring fairness in natural language processing for moral sentiment classification is challenging, particularly under cross-domain shifts where transformer models are increasingly deployed. Using the Moral Foundations Twitter Corpus (MFTC) and Moral Foundations Reddit Corpus (MFRC), this work evaluates BERT and DistilBERT in a multi-label setting with in-domain and cross-domain protocols. Aggregate performance masks important disparities: we observe pronounced asymmetry in transfer, with Twitter$\to$Reddit degrading micro-F1 by 14.9\% versus only 1.5\% for Reddit$\to$Twitter. Per-label analysis reveals fairness violations hidden by overall scores; notably, the \texttt{authority} label exhibits Demographic Parity Differences of 0.22--0.23 and Equalized Odds Differences of 0.40--0.41. To address this gap, we introduce the \emph{Moral Fairness Consistency} (MFC) metric, which quantifies the cross-domain stability of moral foundation detection. MFC shows strong empirical validity, achieving a perfect negative correlation with Demographic Parity Difference ($\rho=-1.000$, $p<0.001$) while remaining independent of standard performance metrics. Across labels, \texttt{loyalty} demonstrates the highest consistency (MFC = 0.96) and \texttt{authority} the lowest (MFC = 0.78). These findings establish MFC as a complementary, diagnosis-oriented metric for fairness-aware evaluation of moral reasoning models, enabling more reliable deployment across heterogeneous linguistic contexts.
\end{abstract}


\end{frontmatter}





\section{Introduction}
\label{sec:introduction}
In recent years, various Transformer-based Language Models (LMs) have been seamlessly integrated into society. The rapid expansion of LM has led to serious concerns about fairness and ethical issues \cite{bansal2022surveybiasfairnessnatural}. For example, large language models (LLMs), which incorporate the ‘self-attention’ mechanism of transformer models \cite{vaswani2017attention}, have been widely adopted in the industry due to their ability to perform various tasks, ranging from simple text processing to complex knowledge generation \cite{naveed2024comprehensiveoverviewlargelanguage}. Today, the latest versions of LLMs are being released by leading tech companies, including GPT-4 by OpenAI \cite{openai2024gpt4technicalreport}, LLaMA-3 by Meta \cite{grattafiori2024llama3herdmodels}, and the newly introduced open-source DeepSeek-V3 \cite{deepseekai2024deepseekv3technicalreport}. The introduction of these models has significantly transformed the economic, academic and geopolitical landscape, marking a new chapter in the artificial intelligence race. The rapid advancements in such models have introduced certain societal risks. Since language models are pre-trained on human language artifacts, often sourced from the internet, various forms of biases and unfair behaviors may be embedded within these models \cite{ali2024understandinginterplayscaledata}. Therefore, open-source pre-trained transformer-based language models such as BERT \cite{devlin2019bertpretrainingdeepbidirectional} continue to serve as important baseline models \cite{rogers2020primerbertologyknowbert} for researching ethical implications. Such open source models provide greater transparency, making them valuable resources \cite{Chen_2024} to further investigate risk mitigation strategies, ensure fairness, and promote ethical AI development in various applications.

To address emerging concerns regarding the fairness of transformer based language models, beyond traditional evaluation metrics for moral sentiment analysis, Moral Foundation Theory (MFT) is employed. This theory, which defines six universal foundations of moral reasoning, is structured into vice and virtue polarities: 
\texttt{Care/Harm}, \texttt{Fairness/Cheating}, \texttt{Loyalty/Betrayal}, \texttt{Authority/\allowbreak Subversion}, \texttt{Purity/Degradation}, and \texttt{Liberty/Oppression} 
\cite{graham2013moral}. One of the primary advantages of vice and virtue polarity in moral sentiment classification is that a broader contextual range across various domains can be captured. For instance, in binary sentiment analysis, only positive and negative sentiments are extracted from text. In contrast, moral sentiment can be identified without reliance on specific linguistic structures, languages, cultural norms, or demographic groups when MFT is used \cite{zangari2024survey}. Therefore, Moral Foundation Theory offers a robust framework for moral sentiment classification and provide a groundwork for researching ethical implications in language models and development of novel metrics for fairness in context of moral sentiment classification. Thus, one of the key reasons why MFT is a suitable framework for moral sentiment classification is its ability to provide a structured moral reasoning framework that is empirically validated across cultures \cite{graham2013moral} which highlights that the six moral foundations operate as universal dimensions across societies, allowing for a wider adoption in Natural Language Processing (NLP) tasks such as in content moderation \cite{khan2025natural}, policy making and conflict resolution \cite{10.1145/3677525.3678694}.

Using Moral Foundation Theory, scholars have generated benchmark datasets for the classification of moral sentiment \cite{twitter, trager2022moral}. Therefore, by applying this universal theory in moral sentiment classification, an attempt can be made to quantify the fairness of language models beyond traditional performance metrics. To address these challenges, this paper will evaluate the extent to which an evaluation framework incorporating novel fairness metrics can bridge a critical research gap in the classification of moral sentiments across various social media platforms.


Therefore, we can formulate a robust \textbf{research question} to address the research gap:
\noindent\textit{To what extent fairness metrics can be developed that account for differences between social media platforms in NLP models to ensure fair classification of moral sentiments?}

Through systematic investigation of this question, this research aims to advance the understanding of fairness in models in the field of morality and provide context to further develop fairness metrics that ensure fair deployment of models.

\section{Related Work}
\label{sec:related_work}
Since language models and their indirect implementations become more ubiquitous, research  regarding the fairness of such implementations become crucial for developing robust models. To grasp this gap in the research, a deeper exploration into existing approaches and relevant theory behind is needed.

As technology advances, the complexity of our problems increases quickly. Especially for sophisticated and subjective tasks concerning morality of any dimension. Yet, researchers like Jiang et al. \cite{jiang2022machineslearnmoralitydelphi}, have started a challenging research in attempting to teach morality in AI systems. This paper proposes an experimental framework called Delphi, that is based on deep neural networks which are capable of reasoning and predicting descriptive ethical judgments that is often aligned with human expectations. The empirical results of this study suggests that Deplhi has a potential to generalize to ethical scenarios with minimal alterations in context and outperforms state of the art neural network models \cite{jiang2022machineslearnmoralitydelphi}. However, despite the good performance of the Delphi framework, the study highlights their key limitations in the reliability. The limitation of the study occurs when the Delphi learns on aggregated statistically dominant behaviors of the data, sometimes pushing 'normative' or 'popular' views, which can be seen as amplification of existing biases. Exploring further the morality in artificially intelligent systems, one of the leading companies in LLM development - Anthropic, have introduced a work regarding the topic of moral self-correction in large language models \cite{ganguli2023capacitymoralselfcorrectionlarge}. This research have found out that language models trained with reinforcement learning from human feedback (RLHF) start to have the ability of morally self-correcting itself at around 22B parameters, which can help them to avoid harmful outputs. Therefore, the results indicate us designing and evaluating language models in terms of morality can be done more precisely in the future. 

Further works in the field of morality in AI systems are grounded by several frameworks in morality, ethics, and values. According to a review of the literature by Vida et al. \cite{vida2023valuesethicsmoralsuse}, the majority of the reviewed works employed Moral Foundation Theory \cite{graham2013moral} as a baseline for their researches, where 49 papers out of 59 with moral frameworks mentioned employed the MFT. However, the MFT is not the sole moral framework used in NLP that focus on morals, ethics and values. Previous studies implement ethics and values through Schwartz's Values Theory and Kohlberg's stages of moral development theory \cite{kiesel-etal-2022-identifying,inproceedings}.

The work of Hoover et al. \cite{twitter} introduces Moral Foundations Twitter Corpus, which remains as a crucial resource for tasks related to moral sentiment analysis, due to high quality hand annotated data for model training and evaluation. Complementary to the Twitter corpus, Trager et al. \cite{trager2022moral}, introduces their own variant of corpus based on the Reddit social media platform. This work also follows the standard based on the MFT theory, extending the existing resource to a fundamentally distinct domain of social media. These important resources have enabled the development of MoralBERT \cite{10.1145/3677525.3678694}, a fined-tuned language model that can be used for classifying moral sentiment in different environments. Extending the BERT family language models for moral sentiment analysis tasks could be linked to the fact that these models show sense of social norms \cite{devlin2019bertpretrainingdeepbidirectional} and demonstrate ability to capture morality by identifying 'right' and 'wrong' \cite{schramowski2022largepretrainedlanguagemodels}. Therefore, MoralBERT utilizes aggregated and domain adversarial training strategies to improve moral sentiment predictions. The fine-tuned model shows notable improvements in single and multi labeled tasks, beating lexicon based methods like Word2Vec in F1 score. Similarly, work of Guo et al. \cite{guo2023datafusionframeworkmultidomain}, expands on multi domain morality learning highlighting the flaw of using heterogeneous data for morality learning by proposing a data fusion framework. This framework allows training on multiple heterogeneous datasets, not just aggregating them, but using a domain adversarial training to align the training data in feature space using a weighted loss function to handle the moral label shift. Therefore, further analysis of fairness metrics is needed for pre-trained models of the BERT family as these models tend to retain biases from the pre-training periods \cite{ali2024understandinginterplayscaledata}.

Despite these advancements in performance for various strategies for moral sentiment classification models, existing approaches still struggle with fairness evaluations \cite{zangari2024survey}. The work of Zangari et al., highlights the results of varying researches with transformer based models, which show how these models over represent moral sentiments from their training data \cite{zangari-etal-2025-me2}. One of the few solutions that improve the explainability of such models is the Chain-of-Though approach introduced by Jin et al., \cite{jin2022makeexceptionsexploringlanguage} to gain insights into the underlying hidden processes of the model. Recent advancements in sentiment analysis problems in terms of fairness focus their work on sensitive attributes such as gender, age, sex and political stance \cite{radaideh2025fairness}. Further research into the fairness of machine learning models indicates that different ML models exhibit varying degrees of fairness when the Equalised Odds metric is applied as a fairness evaluation to the same dataset. \cite{uddin2024novel}. Therefore, existing approaches of fairness evaluation in a context of moral sentiment classification are fundamentally different than simpler sentiment analysis tasks, because the morality concept itself is non-binary \cite{park2024moralitynonbinarybuildingpluralist}. Given this fact, work of Park et al. \cite{park2024moralitynonbinarybuildingpluralist} have proposed a framework where the morality concept is treated pluralistically by building a pluralist moral sentence embedding space via a contrastive learning approach. However, the results indicate that moral pluralism is hard to deduce without supervised approach with human labels.

All these mentioned works are part of a broader research effort to understand machine and algorithm behavior. One of the major works that shed light on the importance of comprehending machine behavior was done by Rahwan et al., where the research highlights the necessity of broad scientific research agenda to study and understand machine behavior not just in computer science discipline, but expanding the research into various scientific fields \cite{article_machine_beh}. One of the motivations for such an interdisciplinary approach is to try to solve the ‘black-box’ nature of algorithms from multiple perspectives, meaning research should combine insights from contrasting fields. For example, computer scientists and engineers working on machine or algorithm behavior cannot fully assess the implications of their work because they lack specialization in evaluating behavioral impacts of their works on a broader scale  \cite{article_machine_beh}. Therefore, in the context of this research, a broader understanding is needed on how language models are evaluated for fairness in such a way that meaningful evaluations are resulted. An extensive empirical comparison of extrinsic fairness metrics in the domain of NLP has been conducted by the work of Czarnowska et al, where the results of this extensive research indicate that various existing fairness metrics are simply parametric variants of the three main fairness metrics proposed \cite{10.1162/tacl_a_00425}, meaning that the fairness is able to be assessed by some of the metrics depending on the setting and the questions being asked by the researchers.

\section{Methodology}
\label{sec:Methodology}

In this section, the general methodology of this research is outlined, which can be divided into three main parts that directly address the research questions: (1) exploratory data analysis and data preparation of two distinct social media datasets to understand moral label distributions across platforms, (2) in-domain and cross-domain model training and evaluation using DistilBERT and BERT models to quantify disparities in moral sentiment classification between Twitter and Reddit domains, and (3) novel fairness metric development and validation specifically designed for moral sentiment classification to measure cross-domain consistency and enable fair evaluation of model behavior across social media platforms. The methodology can be seen in Figure \ref{fig:pipeline_new} of Appendix \ref{sec:apx:first_appendix}.

\subsection{Data}
Two distinct datasets, sourced from relevant social media platforms, are utilized in this research. The datasets, which contain posts and comments from Reddit and Twitter (now X), have been manually annotated in accordance with the Moral Foundation Theory \cite{graham2013moral}. It can be observed that each dataset represents a distinct domain for moral learning, owing to their contrasting characteristics. It is noted that Reddit is characterized by user anonymity and unrestricted character limitations in posts and comments, whereas Twitter is distinguished by reduced anonymity and constrained text capacities.

\subsubsection*{Moral Foundation Twitter Corpus (MFTC) \footnote{\url{https://osf.io/k5n7y/}}}
The dataset comprises of 35,108 'tweets' that were hand labeled by at least three trained annotators, in total 21 distinct annotators contributed to label moral annotations of the MFTC. Seven distinct datasets, each varied by topic, are included in the collection. The MFTC include 'tweets' on the topics from such domains as: BLM, Election, Davidson, Baltimore, MeToo, Sandy and ALM. Originally, the data set was intended to be collected using a Python script connected to the Twitter API. However, due to recent changes in the company's policy, retrieving historical tweets by their ID has become significantly more challenging and costly. As a result, permission was granted to use a version of the dataset which was released by Morteza Dehghani, the author of the MFTC. The dataset, as provided, includes moral labels such as \texttt{Non-Moral}, \texttt{Harm}, \texttt{Cheating}, \texttt{Care}, \texttt{Fairness}, \texttt{Subversion}, 
\texttt{Loyalty}, \texttt{Authority}, \texttt{Degradation}, \texttt{Betrayal}, and \texttt{Purity}. The original MFTC dataset is made availabel in a nested JSON format, thus necessary preprocessing steps such as flattening the structure and transforming are applied to convert into pandas DataFrame format. Through these preprocessing steps, the dataset is transformed into a format containing 33,687 unique 'tweets' and 128,454 total 'tweets,' where individual 'tweets' are annotated by multiple annotators. After preprocessing, the structure of the MFTC dataset appears in the following way as seen in Figure \ref{fig:dataset1}.

\subsubsection*{Moral Foundation Reddit Coprus (MFRC)} This dataset is availabel on Hugging Face\footnote{\url{https://huggingface.co/datasets/USC-MOLA-Lab/MFRC}}, containing 13,995 posts/comments from various 'subreddits' of the Reddit platform. This corpus is divided into three main topic buckets: US Politics with subreddits like 'r/conservative', 'r/antiwork', 'r/politics'; Everyday Morality bucket with subreddits like 'r/IamTheAsshole', 'r/nostalgia', 'r/relationship\_advice' and 'r/confession'; and the last bucket French Politics with subreddits like 'r/geopolitics', 'r/europe', 'r/conservative', 'r/neoliberal', and 'r/worldviews'. Similarly, to the Twitter corpus, moral sentiments were labeled by trained annotators, in total 6 distinct annotators, with an agreement threshold of 50\% for the final labels. The dataset, as provided, includes moral labels such as \texttt{Non-Moral, Thin Morality, Care, Equality, Authority, Proportionality, Loyalty} and \texttt{Purity}. The structure of the MFRC looks in the following way as in Figure \ref{fig:dataset2}.

\subsection{Exploratory Data Analysis}
Before going in-depth with the experimental setup, initial exploratory data analysis has to be performed. During this process, thorough analysis of the existing Moral Foundation Datasets was conducted. This step is considered crucial to understanding the distributions of the moral sentiment labels that are contained in each post or comment. First of all, the Moral Foundation Reddit Corpus is focused on, followed by the Moral Foundation Twitter Corpus.

Figure \ref{fig:allLABELREDDIT} illustrates all different combinations of moral sentiment labels of the MFRC, which were hand annotated by trained expert annotators. First of all, this graph show that label \texttt{Non-Moral} individually occurs the most. The label \texttt{Non-Moral} is assigned to Reddit posts that are neutral and do not exhibit any moral traits. The moral label \texttt{Thin Morality} also shows a high occurrence in the graph, which can be explained by the fact that in MFRC research, before assigning to moral label \texttt{Non-Moral} the annotators should have looked for signs of \texttt{Thin Morality}, where the given text is on the verge of expressing some kind of morality form. To investigate the moral labels, visualizing the unique label distribution could give a detailed overview of the moral label class balance which can be seen in Figure \ref{fig:labelR}. Plotting the graph for each label gives a more narrowed down moral sentiment class distribution. The MFRC contains 8 moral foundation annotations: the most frequent is the label \texttt{Non-Moral} followed by \texttt{Thin Morality}, \texttt{Care}, \texttt{Equality}, \texttt{Authority}, \texttt{Proportionality}, \texttt{Loyalty}, and \texttt{Purity}.

The next step is to compare the Moral Foundation Twitter Corpus with the Reddit corpus, so the cross-domain testing experiments can be designed in such way that moral sentiment label shift can be captured in a controlled way. Therefore, the same visualization approach will be applied on the moral labels of the MFRC. The initial, distribution shows even more mixed annotations compared to the Reddit corpus, where the label \texttt{non-moral} is the most frequently occurring. Thus, to understand the label distribution, plotting the bar chart of the unique label occurrences gives more interpretable insights of the distribution, which can be seen in Figure \ref{fig:twiiterLABE}. Similarly, the label \texttt{non-moral} shows high occurrence, followed with the labels \texttt{harm}, \texttt{cheating}, \texttt{care}, \texttt{fairness}, \texttt{subversion}, \texttt{loyalty}, \texttt{authority}, \texttt{degradation}, \texttt{betrayal}, \texttt{purity} and some none relevant occurrences of \texttt{nm} and \texttt{nh}, which will be removed from the training, validation and evaluation datasets.

\subsection{Data Preparation}
\sloppy
The MFRC dataset includes Reddit comments annotated for the presence of moral language. The annotations follow updated theoretical developments in moral sentiment \cite{trager2022moral}, distinguishing 
\texttt{Proportionality}, \texttt{Equality} as separate moral foundations, rather than subsuming them under a single \texttt{Fairness} label as in traditional MFT schemas. This approach results in 
\texttt{Non-Moral}, \texttt{Thin Morality}, \texttt{Care}, \texttt{Equality}, \texttt{Authority}, \texttt{Proportionality}, and \texttt{Loyalty}. 
For the purposes of model consistency across datasets, we map \texttt{Equality} and \texttt{Proportionality} labels to \texttt{Fairness}, aligning with the broader MFT based taxonomy and enabling cross dataset comparison. This label unification is necessary since the MFTC does not distinguish between \texttt{Equality} and \texttt{Proportionality}, similar mapping was applied in the work of MoralBERT \cite{10.1145/3677525.3678694} to fine-tune a BERT model with increased moral awareness.

\begin{figure*}[t]
  \Description{Label distribution.}
  \centering
  \subfloat[\textbf{MFTC Label Distribution}\label{fig:mftc_dist}]{
    \includegraphics[width=0.45\textwidth]{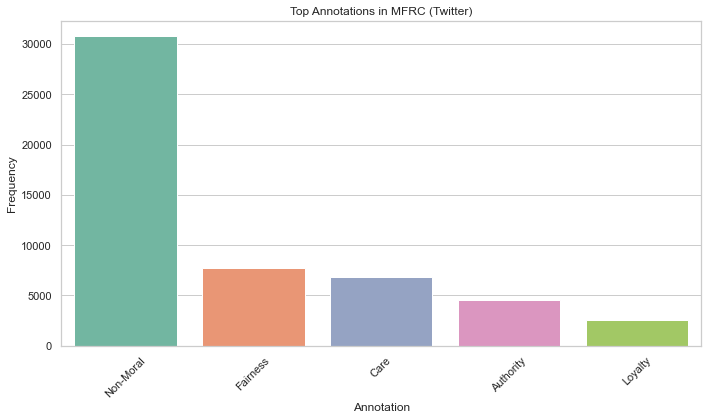}
  }\hfill
  \subfloat[\textbf{MFRC Label Distribution}\label{fig:mfrc_dist}]{
    \includegraphics[width=0.45\textwidth]{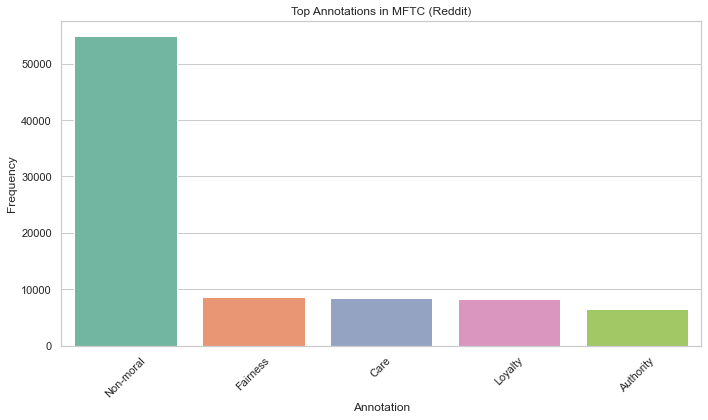}
  }
  \caption{Comparison of label distributions in the two corpora after label harmonization to 5 morality labels.}
  \label{fig:label_dists}
\end{figure*}

The MFTC provides tweets annotated with a broader set of moral labels, which includes both virtues and vices. The annotated categories include:
\texttt{Non-Moral}, \texttt{Care}, \texttt{Fairness}, \texttt{Loyalty}, \texttt{Authority}, \texttt{Purity}, \texttt{Harm}, \texttt{Cheating}, \texttt{Subversion}, \texttt{Betrayal}, and \texttt{Degradation}.

To ensure interoperability between the datasets during training and evaluation, we perform label alignment by focusing on the following shared categories across corpora:
\texttt{Care, Fairness, Loyalty, Authority, Non-Moral}. Tweets or Reddit comments with labels not covered in both corpora are excluded from in-domain and cross-domain training and evaluation to preserve label compatibility. Additionally, regular data cleaning methods for textual data, such as lower casing, removing whitespace and special characters were applied to the whole corpus of the MFTC and MFRC. No lemmatization or stemming was applied, as transformer models are designed to process subword units and learn contextual representations inherently \cite{wang-etal-2022-simkgc}. Excessive preprocessing might undermine this contextual richness, which can have strong influence on morality learning tasks. The final moral label distribution, used for the experiments can be seen from the Figure \ref{fig:label_dists}.

\subsection{Model}
\textbf{B}idirectional \textbf{E}ncoder \textbf{R}epresentations from \textbf{T}ransformers or in short BERT is a pre-trained deep learning model for NLP tasks. This model is built upon the Transformer architecture \cite{vaswani2017attention}, which allows the model to leverage self-attention mechanism to capture contextual word representations bidirectionally \cite{devlin2019bertpretrainingdeepbidirectional}. This means that the BERT can process text from both directions simultaneously. The ability to construct deeper context aware word embeddings is a crucial advantage \cite{endalie2025fine} of using such models for multi-label classification problem of moral sentiment analysis.

\begin{equation}
\text{Attention}(Q, K, V) = \text{softmax} \left( \frac{QK^T}{\sqrt{d_k}} \right) V
\label{eq:attention}
\end{equation}

In this research, BERT and DistilBERT models are adopted for multi-labeled moral sentiment classification task to assess the extent on which we can employ new fairness metric. This choice can be rationalized due to key advantages of using such model for multi-label classification tasks on moral sentiment labels. Moral sentiments can be highly contextual and unique for specific social media platforms. Therefore, BERTs ability to handle deeper contextual understanding of the textual data can be helpful for moral sentiment analysis tasks \cite{schramowski2022largepretrainedlanguagemodels}. Using BERT family language models for moral sentiment analysis tasks could be linked to the fact that these models show a sense of social norms \cite{devlin2019bertpretrainingdeepbidirectional} and demonstrate the ability to capture morality by identifying \textbf{"right"} and \textbf{"wrong"} \cite{schramowski2022largepretrainedlanguagemodels}.

\subsection{Loss Function}

Binary Cross-Entropy with Logits Loss is a deliberate choice for multi-label classification tasks where each instance may belong to zero or more classes. While more advanced techniques like Focal Loss \cite{10947681} exist that improve performance under imbalanced data, the main target of this paper was to study fairness behavior rather than trying to maximize the performance. Therefore, in moral sentiment classification, the BCE loss is appropriate because it allows for independent probabilistic modeling of each moral dimension \cite{tiwari2025advancingvulnerabilityclassificationbert}. 
BCEWithLogitsLoss is defined as:

\begin{equation}
\mathcal{L}_{\text{BCELogits}}(\mathbf{y}, \mathbf{z}) = - \frac{1}{L} \sum_{i=1}^{L} \left[ y_i \log \sigma(z_i) + (1 - y_i) \log (1 - \sigma(z_i)) \right]
\end{equation}

where $\sigma(z_i) = \frac{1}{1 + e^{-z_i}}$ is the sigmoid activation applied to the raw logits $z_i$, $y_i \in \{0,1\}$ is the true label, and $L$ is the number of labels.

\subsection{Model Training}
For the models, DistilBERT \cite{sanh2020distilbertdistilledversionbert} and full BERT architectures are fine-tuned using the Transformers library by HuggingFace.

DistilBERT, a distilled version of BERT, serves as our lightweight baseline due to its reduced size and faster training time, making it an efficient benchmark for real-world applications. It is approximately 40\% smaller and 60\% faster than BERT-base while retaining 95\% of its performance on standard benchmarks \cite{sanh2020distilbertdistilledversionbert}. Its efficiency makes it ideal for understanding baseline behavior and assessing performance across domains. 
We additionally experiment with the full BERT model to serve as a comparative upper baseline. While computationally heavier, BERT offers a higher representational capacity, potentially leading to better generalization in complex, multi-label tasks such as cross-domain moral sentiment analysis. The full BERT model was incorporated to assess the trade-off between performance and computational complexity. Each model was fine-tuned separately on the Moral Foundations Reddit Corpus and Moral Foundations Twitter Corpus.

The classes such as \texttt{DistilBertForSequenceClassification} and \texttt{BertForSequenceClassification} are utilized from HuggingFace, setting num\_labels according to the number of labels after label harmonization, which results in 5 moral labels. For the MFRC, we treated the task as multi-label classification, following the dataset’s allowance of multiple moral foundations per post. For MFTC, which uses one label per tweet, we employed multi-label classification. The problem type was set accordingly using HuggingFace’s problem\_type flag for the multi-label case. For in-domain experiments, a 80/10/10 split is used for training, validation, and testing. In cross-domain scenarios, the entire source domain dataset is used for fine-tuning, while testing is conducted on the target domain.

This paper utilizes AdamW optimizer with a learning rate of 2e-5, which is a widely accepted setting for fine-tuning transformer models for both BERT and DistilBERT. The models were trained for five epochs on a single NVIDIA A-100 GPU (via Google Colab), binary cross-entropy with logits loss for multi-label tasks, automatically handled by the Transformers framework. To ensure reproducibility, we fixed the random seed before training and did not apply any oversampling or balancing techniques, instead preserving the natural label distribution to reflect real world class imbalances. This decision was deliberate, aiming to evaluate model robustness on underrepresented moral categories, especially relevant for cross-domain evaluation, as correcting for class or label balance is not always necessary and often might bring hidden risks \cite{carriero2024harmsclassimbalancecorrections}. However, it is important to note that transformer models are highly sensitive to hyperparameter configurations where performance and fairness can be directly influenced by the configuration of the learning rate, batch size and number of epochs \cite{devlin2019bertpretrainingdeepbidirectional}.

\subsubsection*{Multi-labeled Classification Task for Moral Sentiment Labels (In-domain vs. Cross-domain)}
After training the models, the performance should be assessed under two conditions:
\begin{itemize}
    \item \textbf{In-domain}: the model was fine-tuned and tested in the same social media domain, e.g train and tested for moral sentiment classification on MFTC.
    \item \textbf{Cross-domain}: the model fine-tuned on one social media corpus is tested on a different social media domain full data.
\end{itemize}

These steps are critical to understand how moral sentiment classification models behave across domains.

\subsection{Evaluation: Performance and Fairness Metrics for Moral Sentiment Classification}

We evaluate each model both in-domain (Reddit on Reddit, Twitter on Twitter) and cross-domain (Reddit on Twitter, Twitter on Reddit). This allows us to test generalizability, domain robustness, and transferability of learned moral semantics. Evaluation metrics include BCE loss value, micro-F1 score, and Exact Match Ratio (EMR). These metrics are chosen to reflect performance at both instance and moral label levels. For more granular evaluation per label F1, Precision and Recall is computed for in-domain and cross-domain scenarios. 
Additionally, we use bootstrapping (n = 1000) to compute 95\% confidence intervals (CI) for all performance metrics, which provides robust estimates of variability and statistical reliability. This is particularly important in multi-label moral classification, where label sparsity and domain shifts can introduce high variance in model performance. 

To assess model fairness, we also compute per label Demographic Parity Difference (DP)\footnote{\url{https://fairlearn.org/main/api_reference/generated/fairlearn.metrics.demographic_parity_difference.html}} and per label Equalised Odds Difference (EO)\footnote{\url{https://fairlearn.org/main/api_reference/generated/fairlearn.metrics.equalized_odds_difference.html}} as fairness metrics. In our setup, the sensitive attribute is the platform origin: \textbf{Reddit} and \textbf{Twitter}, which are treated as a binary group attribute. These metrics are computed using the output logits and prediction files from the cross-domain experiments. This setup allows us to measure fairness violations across domains, ensuring that a models moral label predictions do not systematically differ based on the source domain alone. Similarly, we use bootstrapping to compute 95\% confidence interval for robust analysis, where n = 1000.

\subsubsection*{Novel Metric Exploration and Validation}
To evaluate the validity of the proposed new fairness metric in the context of the transformer models for moral sentiment classification, examining the empirical association with existing fairness and performance metrics for cross-domain fine-tuning scenarios. This paper will employ the Spearman correlation coefficient \cite{sheskin2003handbook}, which is an appropriate choice given the non-linear nature of the moral foundations relationship in the context of deep learning. 

\section{Results}
\label{sec:Results}
This section presents and highlights the results obtained from the conducted experiments of fine-tuning DistilBERT and BERT models for the multi-label classification of moral foundation labels. The experiments can be divided into two categories: in-domain and cross-domain. For in-domain scenarios, a model is fine-tuned on one social media domain and tested on the same domain, for example fine-tuned on Reddit and tested on Reddit for multi-label classification. This results in four scenarios of experiments for in-domain: two on DistillBERT and two for full BERT models. Similarly, cross-domain experiments maintain the same logic. The model is fine-tuned on one social media domain and tested on different social media data for multi-label classification task. Therefore, the cross-domain approach results in four 
experiment scenarios. For the traditional performance metrics this section utilizes Binary Cross Entropy with Logits as loss function, micro-F1 score and Exact Match Ratio (EMR). Furthermore, the per label performance results are displayed against the label sparsity and F1 score as seen in Figure \ref{fig:label_sparc} of Appendix \ref{sec:apx:first_appendix}.

\subsection{In-domain Results}

The in-domain results can be observed from Table \ref{tab:in_domain_results} and Figure \ref{fig:perf_deg}. From the results of the experiments several key observations can be seen. Transformer models that are trained on MFTC outperformed those models that were trained on MFRC dataset. For instance, the lighter and faster DistilBERT model of in-domain scenario achieved 0.772 micro-F1 score and Exact Match Ratio of 0.742, compared to the micro-F1 score of 0.687 and EMR of 0.645 for the DistilBERT fine-tuned on MFRC dataset. The loss of models that used MFTC is also noticeably lower than the models that used MFRC for fine-tuning, where loss 0.372 for MFRC and loss of 0.286 for the MFTC scenarios where DistilBERT was used. Similar trend also shows for the BERT in-domain scenarios, where the MFTC trained models have micro-F1 of 0.768 and for MFRC trained model the micro-F1 score is 0.685. The MFRC trained BERT model shows higher loss of 0.379 and lower EMR of 0.640, with comparison to the MFTC trained BERT model gaining lower loss of 0.291 and higher EMR of 0.737, showing better classification results.

\begin{table*}[t]
  \caption{Overall Model Performance Results In-domain vs.\ Cross-domain}
  \label{tab:overall_eval}
  \centering

  \subfloat[\textbf{In-domain experiments}\label{tab:in_domain_results}]{
    \begin{minipage}{0.48\textwidth}
      \centering
      \small
      \begin{tabular}{lccc}
        \toprule
        \textbf{Scenario} & \textbf{Loss} & \textbf{Micro-F1} & \textbf{Accuracy/EMR}\\
        \midrule
        DistilBERT MFRC $\to$ MFRC & 0.372 & 0.687 & 0.645\\
        DistilBERT MFTC $\to$ MFTC & 0.286 & 0.772 & 0.742\\
        \midrule
        BERT  MFRC $\to$ MFRC       & 0.379 & 0.685 & 0.640\\
        BERT  MFTC $\to$ MFTC       & 0.291 & 0.768 & 0.737\\
        \bottomrule
      \end{tabular}
    \end{minipage}
  }
  \hfill
  \subfloat[\textbf{Cross-domain experiments}\label{tab:cross_domain_results}]{
    \begin{minipage}{0.48\textwidth}
      \centering
      \small
      \begin{tabular}{lccc}
        \toprule
        \textbf{Scenario} & \textbf{Loss} & \textbf{Micro-F1} & \textbf{Accuracy/EMR}\\
        \midrule
        DistilBERT MFRC $\to$ MFTC & 0.421 & 0.672 & 0.650\\
        DistilBERT MFTC $\to$ MFRC & 0.688 & 0.623 & 0.643\\
        \midrule
        BERT  MFRC $\to$ MFTC       & 0.416 & 0.673 & 0.649\\
        BERT  MFTC $\to$ MFRC       & 0.648 & 0.624 & 0.643\\
        \bottomrule
      \end{tabular}
    \end{minipage}
  }
\end{table*}

\subsection{Cross-domain Results}
The results of the cross-domain experiments can be seen from Table \ref{tab:cross_domain_results} and Figure \ref{fig:perf_deg}, where the models were fine-tuned on one social media data set and tested on different social media data. As expected, the cross-domain experiments revealed slightly different performance metrics. Applying the models to a different domain decreased the overall performance. This performance degradation can be highly visible from the increasing loss metric for all of the scenarios and reduced micro-F1 score. For instance, DistilBERT fine-tuned on MFTC and tested on MFRC shows significant drop in micro-F1 score, going from 0.772 for in-domain settings to 0.623 for cross-domain scenarios. Similar pattern can be observed from the other experiment scenarios, for example the BERT fine-tuned on MFTC, the performance in terms of micro-F1 have dropped from 0.768 to 0.624. This performance of morality classification ability drops when the model changes domain, and more interestingly, it was not a symmetric performance drop. Models trained on the MFTC and tested on the MFRC have shown bigger drop in performance. For instance, for DistilBERT the micro-F1 decreased by 0.14, and the EMR dropped by 0.09 units. On contrast, models fine-tuned on MFRC and evaluated on MFTC data did not show massive loss in performance for the micro-F1 score, only decreasing by 0.015 units for the cross-domain experiment using DistilBERT. 

Interestingly, the EMR for the MFRC $\rightarrow$ MFTC models have improved slightly compared to the in-domain results. For DistilBERT, the EMR increased from 0.645 (MFRC $\rightarrow$ MFRC) to 0.650 (MFRC $\rightarrow$ MFTC). A similar pattern was observed for BERT, with EMR increasing from 0.640 to 0.649. This suggests that while individual label predictions were slightly less accurate on average, the models more frequently predicted the exact complete set of labels correctly when transferring from MFRC to MFTC compared to their performance within the MFRC domain.

\begin{table*}[t]
  \caption{Per-label F$_1$ (95\% CI) for In-domain vs.\ Cross-domain}
  \label{tab:f1_fullwidth}
  \centering
  \scriptsize
  \setlength{\tabcolsep}{4pt}
  \renewcommand{\arraystretch}{1.05}

  \subfloat[\textbf{In-domain experiments}\label{tab:f1_in}]{
    \begin{minipage}{0.49\textwidth}
      \centering
      \begin{tabularx}{\linewidth}{lYYYY}
        \toprule
              & \multicolumn{2}{c}{DistilBERT} & \multicolumn{2}{c}{BERT} \\
        Label & MFRC$\to$MFRC & MFTC$\to$MFTC & MFRC$\to$MFRC & MFTC$\to$MFTC \\
        \midrule
        authority & 0.38\,(0.34–0.42) & 0.45\,(0.41–0.48) & 0.38\,(0.34–0.42) & 0.45\,(0.40–0.48) \\
        care      & 0.56\,(0.53–0.59) & 0.58\,(0.56–0.61) & 0.57\,(0.54–0.59) & 0.58\,(0.55–0.61) \\
        fairness  & 0.50\,(0.46–0.53) & 0.64\,(0.62–0.67) & 0.51\,(0.48–0.54) & 0.64\,(0.61–0.66) \\
        loyalty   & 0.41\,(0.35–0.47) & 0.57\,(0.54–0.59) & 0.39\,(0.34–0.45) & 0.56\,(0.53–0.58) \\
        non-moral & 0.82\,(0.81–0.82) & 0.88\,(0.87–0.89) & 0.82\,(0.81–0.83) & 0.88\,(0.87–0.88) \\
        \bottomrule
      \end{tabularx}
    \end{minipage}
  }
  \hfill
  \subfloat[\textbf{Cross-domain experiments}\label{tab:f1_cross}]{
    \begin{minipage}{0.49\textwidth}
      \centering
      \begin{tabularx}{\linewidth}{lYYYY}
        \toprule
              & \multicolumn{2}{c}{DistilBERT} & \multicolumn{2}{c}{BERT} \\
        Label & MFRC$\to$MFTC & MFTC$\to$MFRC & MFRC$\to$MFTC & MFTC$\to$MFRC \\
        \midrule
        authority & 0.30\,(0.28–0.31) & 0.03\,(0.03–0.04) & 0.31\,(0.30–0.32) & 0.04\,(0.03–0.04) \\
        care      & 0.42\,(0.41–0.43) & 0.08\,(0.07–0.09) & 0.46\,(0.45–0.47) & 0.09\,(0.08–0.10) \\
        fairness  & 0.34\,(0.33–0.35) & 0.10\,(0.09–0.11) & 0.36\,(0.35–0.37) & 0.12\,(0.11–0.13) \\
        loyalty   & 0.40\,(0.39–0.41) & 0.06\,(0.05–0.08) & 0.38\,(0.37–0.39) & 0.07\,(0.06–0.08) \\
        non-moral & 0.81\,(0.80–0.81) & 0.79\,(0.79–0.79) & 0.81\,(0.80–0.81) & 0.79\,(0.79–0.79) \\
        \bottomrule
      \end{tabularx}
    \end{minipage}
  }
\end{table*}

\begin{table*}[t]
  \caption{Per-label Recall (95\% CI) for In-domain vs.\ Cross-domain}
  \label{tab:recall_fullwidth}
  \centering
  \scriptsize
  \setlength{\tabcolsep}{4pt}
  \renewcommand{\arraystretch}{1.05}

  \subfloat[\textbf{In-domain experiments}\label{tab:recall_in}]{
    \begin{minipage}{0.49\textwidth}
      \centering
      \begin{tabularx}{\linewidth}{lYYYY}
        \toprule
              & \multicolumn{2}{c}{DistilBERT} & \multicolumn{2}{c}{BERT} \\
        Label & MFRC$\to$MFRC & MFTC$\to$MFTC & MFRC$\to$MFRC & MFTC$\to$MFTC \\
        \midrule
        authority & 0.34\,(0.30–0.38) & 0.38\,(0.34–0.42) & 0.37\,(0.32–0.41) & 0.38\,(0.34–0.42) \\
        care      & 0.53\,(0.50–0.57) & 0.54\,(0.51–0.58) & 0.56\,(0.53–0.60) & 0.54\,(0.51–0.58) \\
        fairness  & 0.46\,(0.42–0.49) & 0.62\,(0.59–0.65) & 0.48\,(0.44–0.52) & 0.63\,(0.59–0.66) \\
        loyalty   & 0.34\,(0.28–0.40) & 0.53\,(0.50–0.57) & 0.34\,(0.28–0.40) & 0.54\,(0.51–0.57) \\
        non-moral & 0.80\,(0.79–0.81) & 0.87\,(0.86–0.88) & 0.80\,(0.78–0.81) & 0.87\,(0.86–0.88) \\
        \bottomrule
      \end{tabularx}
    \end{minipage}
  }
  \hfill
  \subfloat[\textbf{Cross-domain experiments}\label{tab:recall_cross}]{
    \begin{minipage}{0.49\textwidth}
      \centering
      \begin{tabularx}{\linewidth}{lYYYY}
        \toprule
              & \multicolumn{2}{c}{DistilBERT} & \multicolumn{2}{c}{BERT} \\
        Label & MFRC$\to$MFTC & MFTC$\to$MFRC & MFRC$\to$MFTC & MFTC$\to$MFRC \\
        \midrule
        authority & 0.21\,(0.20–0.22) & 0.01\,(0.01–0.02) & 0.23\,(0.22–0.24) & 0.01\,(0.01–0.02) \\
        care      & 0.38\,(0.37–0.39) & 0.04\,(0.03–0.04) & 0.45\,(0.44–0.46) & 0.04\,(0.04–0.05) \\
        fairness  & 0.27\,(0.26–0.28) & 0.05\,(0.05–0.06) & 0.30\,(0.29–0.31) & 0.06\,(0.06–0.07) \\
        loyalty   & 0.29\,(0.28–0.30) & 0.03\,(0.02–0.04) & 0.28\,(0.27–0.29) & 0.03\,(0.03–0.04) \\
        non-moral & 0.84\,(0.84–0.85) & 0.98\,(0.98–0.98) & 0.84\,(0.83–0.84) & 0.98\,(0.98–0.98) \\
        \bottomrule
      \end{tabularx}
    \end{minipage}
  }
\end{table*}

\begin{table*}[t]
  \caption{Per-label Precision (95\% CI) for In-domain vs.\ Cross-domain}
  \label{tab:precision_fullwidth}
  \centering
  \scriptsize
  \setlength{\tabcolsep}{4pt}
  \renewcommand{\arraystretch}{1.05}

  \subfloat[\textbf{In-domain experiments}\label{tab:prec_in}]{
    \begin{minipage}{0.49\textwidth}
      \centering
      \begin{tabularx}{\linewidth}{lYYYY}
        \toprule
              & \multicolumn{2}{c}{DistilBERT} & \multicolumn{2}{c}{BERT} \\
        Label & MFRC$\to$MFRC & MFTC$\to$MFTC & MFRC$\to$MFRC & MFTC$\to$MFTC \\
        \midrule
        authority & 0.42\,(0.37–0.46) & 0.53\,(0.49–0.58) & 0.39\,(0.35–0.43) & 0.52\,(0.47–0.56) \\
        care      & 0.57\,(0.54–0.60) & 0.62\,(0.59–0.65) & 0.56\,(0.53–0.59) & 0.62\,(0.59–0.65) \\
        fairness  & 0.54\,(0.50–0.57) & 0.65\,(0.63–0.68) & 0.53\,(0.50–0.56) & 0.64\,(0.61–0.67) \\
        loyalty   & 0.50\,(0.44–0.56) & 0.59\,(0.56–0.62) & 0.45\,(0.39–0.52) & 0.56\,(0.53–0.59) \\
        non-moral & 0.82\,(0.81–0.83) & 0.88\,(0.87–0.88) & 0.83\,(0.82–0.84) & 0.87\,(0.87–0.88) \\
        \bottomrule
      \end{tabularx}
    \end{minipage}
  }
  \hfill
  \subfloat[\textbf{Cross-domain experiments}\label{tab:prec_cross}]{
    \begin{minipage}{0.49\textwidth}
      \centering
      \begin{tabularx}{\linewidth}{lYYYY}
        \toprule
              & \multicolumn{2}{c}{DistilBERT} & \multicolumn{2}{c}{BERT} \\
        Label & MFRC$\to$MFTC & MFTC$\to$MFRC & MFRC$\to$MFTC & MFTC$\to$MFRC \\
        \midrule
        authority & 0.45\,(0.44–0.47) & 0.43\,(0.36–0.50) & 0.44\,(0.42–0.45) & 0.46\,(0.39–0.54) \\
        care      & 0.45\,(0.44–0.46) & 0.79\,(0.75–0.83) & 0.46\,(0.45–0.47) & 0.77\,(0.73–0.81) \\
        fairness  & 0.44\,(0.43–0.45) & 0.55\,(0.51–0.58) & 0.43\,(0.42–0.44) & 0.54\,(0.51–0.57) \\
        loyalty   & 0.62\,(0.61–0.63) & 0.50\,(0.42–0.58) & 0.60\,(0.58–0.61) & 0.45\,(0.39–0.53) \\
        non-moral & 0.76\,(0.76–0.77) & 0.65\,(0.65–0.65) & 0.77\,(0.77–0.77) & 0.65\,(0.65–0.66) \\
        \bottomrule
      \end{tabularx}
    \end{minipage}
  }
\end{table*}

Table \ref{tab:f1_fullwidth}, Table \ref{tab:recall_fullwidth} and Table \ref{tab:precision_fullwidth} presents the F\textsubscript{1}, Recall and Precision scores for each moral label with confidence interval (CI) of 95\% with bootstrap of 1000 samples, to get more granular results per label and for cross-domain and in-domain result interpretations. 

The label \texttt{non-moral} row is inspected first, as it is the majority label for both of the social media domains: both models achieved F\textsubscript{1} > 0.80 when they were evaluated in-domain and their scores fell by less than 0.02 after the domain switch. The result is expected because 'non-moral' instances constitute the majority label in both Reddit and Twitter; the models therefore encountered thousands of neutral examples during training and distinguish 'non-moral' instances with good accuracy for in-domain and cross-domain scenarios. Stable recall score for cross and in-domain and slight drop in precision, indicate that models are still finding almost all of the true positives, but making more mislabeling in the cross-domain cases. 

For the \texttt{loyalty} label, scoring approximately 0.57 for in-domain Twitter case and 0.41 for in-domain Reddit case. However, the F\textsubscript{1} score drops severely to 0.06 for cross-domain scenarios, where the MFTC fine-tuned model gets evaluated on MFRC dataset. This drop is similar for DistilBERT and BERT models. On contrast, the DistilBERT and BERT fine-tuned on MFRC did not experience such drastic F\textsubscript{1} score drop for the \texttt{loyalty} label. Reddits longer and richer language might have trained the model to retain much more clear distinction for the \texttt{loyalty} label. When the model is trained on Reddit and tested on Twitter precision actually rises to 0.62, but recall goes down to 0.28 for DistilBERT and BERT approximately the same. Learned attributes for the label \texttt{loyalty} from Reddit generalise only partially to Twitter, thus the model keeps spotting many true positives but over predicts instances that resembles false positives on Twitter.

Similar observation can be seen for \texttt{fairness} moral label. In-domain DistilBERT and BERT models for MFRC show F\textsubscript{1} score of 0.50 and 0.51, with confidence intervals of 0.46–0.53 and 0.48–0.54, respectively. However, for cross-domain cases the scores drop to 0.34 (0.33–0.35) for DistilBERT and 0.36 (0.35–0.37) for BERT, confirming that \texttt{fairness} cues do not transfer well across domains. When the MFRC models are evaluated on Twitter the precision is 0.44 while recall goes to 0.29, indicating that many Reddit fairness cues are still detected but mislabeled in the noisier Twitter domain. The inverse illustrates a similar pattern as previous labels, where the Twitter fine-tuned models evaluated on Reddit retain precision around 0.55, but the recall drops below 0.06.

The in-domain cases reveal steady performance for the \texttt{care} label. Both DistilBERT and BERT models perform similarly with F\textsubscript{1} scores steadily showing 0.56 and 0.58, with slight overlapping in confidence intervals. In contrast, the cross-domain cases show similar asymmetry as other labels in transfer performance. For models fine-tuned on Reddit and evaluated on Twitter, the F\textsubscript{1} score falls from 0.56 to 0.42 (CI 0.41–0.43) for DistilBERT and from 0.57 to 0.46 (0.45–0.47) for BERT. When compared to the confidence intervals of the in-domain cases, the intervals no longer overlap which indicate a true loss of performance. The models fine-tuned on the MFTC and evaluated on Reddit, illustrate the same F\textsubscript{1} score loss going from 0.58 to 0.08 (CI 0.07-0.09) for DistilBERT and to 0.09 (CI 0.08-0.10) for BERT models. For in-domain cases, the precision and recall is consistent for both of the BERT and DistilBERT models. For MFRC $\rightarrow$ MFTC scenarios, the recall slightly drops to 0.38 and 0.45 for DistilBERT and BERT models. However, when reversed, the recall goes down dramatically to 0.04, meaning that the Twitter fine-tuned models mislabel almost every statement that might contain cues for \texttt{care} moral foundation.

Lastly, the \texttt{authority} label shows the same pattern as the other labels. Where the in-domain F\textsubscript{1} scores sit at 0.38 (0.34–0.42) on Reddit fine-tuned models, and 0.45 (0.41–0.48) on Twitter fine-tuned models. When the MFRC models are tested on MFTC, F\textsubscript{1} score drops to 0.30 and 0.31 for DistilBERT and BERT, and the confidence intervals (0.28–0.32) remain well below the in-domain Twitter interval, signaling a consistent degradation. For in-domain cases, the precision is moderate 0.42 and 0.53 for DistillBERT and approximately similar for the BERT model, although in the latter model Reddit model is the worst performed in precision. For cross-domain cases, the Reddit fine-tuned DistilBERT model maintains the precision at 0.44, but the recall goes down to 0.21, confirming false negatives that shows degradation in \texttt{authority} label classification. For the Twitter fine-tuned models precision got slight drop, but the recall shrinks almost to 0.01-0.02, showing severe mislabeling and no cues for detecting \texttt{authority} moral label outside the training domain.

\subsection{Fairness Analysis}
From the previous results, we conclude that in multi-label moral foundation classification, the reliance on a single aggregate metric proves insufficient, as it inherently masks critical biases that exhibit uneven distribution across distinct moral dimensions, therefore for the fairness metrics computing per label metrics will reveal granular results that will be useful for developing a novel metric that account morality foundations. The two social media platforms serve as our sensitive group attributes for the fairness metrics, due to our cross-domain experiments. This research utilizes Fairlearn, an open source library, due to its well tested \cite{bird2020fairlearn}, standard ways to measure fairness and giving overall and group specific rates or error gaps.

In our multi‐label moral label classification setting, aggregate parity metrics conceal label specific imbalances. Each moral foundation has distinct normative weight and frequency on Twitter versus Reddit. For example, \texttt{loyalty} may be rare on Twitter but central in certain subreddits. To highlight these differences, we compute per label Demographic Parity Difference and Equalized Odds Difference for each moral dimension \(m\). This metric captures the models ability to correctly detect the True Positive Rate (TRP) and minimising False Positives Rate (FPR) for each moral foundation is equally reliable across domains.

\begin{align}
\Delta_{\mathrm{DP}}^{(m)}
&= 
P\bigl(\hat y_{m} = 1 \mid A = \mathrm{Twitter}\bigr)
- P\bigl(\hat y_{m} = 1 \mid A = \mathrm{Reddit}\bigr),
\\[8pt]
\Delta_{\mathrm{EO}}^{(m)}
&=
\max\!\bigl\{
\lvert \mathrm{TPR}_{\mathrm{Twitter}}^{(m)} - \mathrm{TPR}_{\mathrm{Reddit}}^{(m)}\rvert,\;
\lvert \mathrm{FPR}_{\mathrm{Twitter}}^{(m)} - \mathrm{FPR}_{\mathrm{Reddit}}^{(m)}\rvert
\bigr\},
\end{align}

where for each domain \(d\in\{\mathrm{Twitter},\mathrm{Reddit}\}\):

\begin{align}
\mathrm{TPR}_{d}^{(m)}
&= 
P\bigl(\hat y_{m} = 1 \mid y_{m} = 1,\; A = d\bigr),
\\[4pt]
\mathrm{FPR}_{d}^{(m)}
&= 
P\bigl(\hat y_{m} = 1 \mid y_{m} = 0,\; A = d\bigr).
\end{align}

Table \ref{tab:fairness_metrics_multirow} presents the fairness metrics for both DistilBERT and BERT models across all moral foundation labels. The results revealed significant cross-domain disparities, with \texttt{authority} showing the largest demographic parity differences for both models DistilBERT 0.22 (0.22–0.23) and BERT 0.23 (0.22–0.23). Similarly, \texttt{authority} exhibited the highest equalized odds differences DistilBERT 0.40 (0.39–0.41) and BERT 0.41 (0.40–0.42), indicating substantial disparities in both prediction rates and error patterns across Twitter and Reddit platforms. In contrast, \texttt{loyalty} demonstrated the smallest demographic parity gaps for both models DistilBERT 0.03 (0.03–0.03) and BERT 0.04 (0.04–0.04), though its equalized odds differences remained substantial at 0.20 and 0.22 respectively. These per label findings demonstrated that moral themes with lower base rates, such as \texttt{authority}, were disproportionately affected by cross-domain biases despite their significance. For the label \texttt{care} demographic parity difference is 0.04 (0.04–0.04) for both DistilBERT and BERT, with equalized odds differences of 0.25 (0.24-0.26) and 0.24 (0.23-0.25), indicating a moderate change in how \texttt{care} is detected on Twitter versus Reddit. Demographic parity difference score for the label \texttt{fairness} is 0.05 (0.05–0.05) for DistilBERT and 0.06 (0.05–0.06) for BERT with equalised odds difference of 0.23 (0.22–0.24) and 0.22 (0.21–0.23). This suggests some domain bias but less severe than for moral label of \texttt{authority}.

\begin{table} 
  \centering
  \caption{Cross-Domain Fairness Metrics by Model and Label} 
  \label{tab:fairness_metrics_multirow}
  \scriptsize 
  \setlength{\tabcolsep}{4pt} 
  \renewcommand{\arraystretch}{1.05} 
  \begin{tabular}{l l c c}
    \toprule
    Model & Label &  $\Delta$ Demographic Parity & $\Delta$ Equalised Odds \\
    \midrule
    \multirow{5}{*}{DistilBERT} 
     & authority & 0.22\,(0.22-0.23) & 0.40\,(0.39-0.41) \\
     & care & 0.04\,(0.04-0.05) & 0.26\,(0.25-0.28) \\
     & fairness & 0.05\,(0.05-0.05) & 0.22\,(0.21-0.23) \\
     & loyalty & 0.03\,(0.03-0.03) & 0.20\,(0.19-0.21) \\
     & non-moral & 0.08\,(0.08-0.08) & 0.34\,(0.33-0.36) \\
    \midrule 
    \multirow{5}{*}{BERT} 
     & authority & 0.23\,(0.22-0.23) & 0.41\,(0.40-0.42) \\
     & care & 0.04\,(0.04-0.04) & 0.24\,(0.23-0.26) \\
     & fairness & 0.05\,(0.05-0.06) & 0.24\,(0.23-0.25) \\
     & loyalty & 0.04\,(0.04-0.04) & 0.22\,(0.21-0.23) \\
     & non-moral & 0.09\,(0.09-0.09) & 0.41\,(0.40-0.42) \\
    \bottomrule
  \end{tabular}
\end{table}

Together, these per label results reveal a clear ordering of cross-domain robustness, where \texttt{loyalty} is most stable, \texttt{care} and \texttt{fairness} exhibit moderate bias, and \texttt{authority} alongside \texttt{non-moral} suffer the greatest disparities. This pattern highlights, that aggregate fairness measures would mask critical, label specific weaknesses. Particularly for moral dimensions with low occurrences, but with high moral weight. To ensure fair moral foundation tagging across social platforms, proposing a fairness metric based on the previous results would highlight the potentials in this research area.

\subsection{Novel Fairness Metric Exploration}

Existing group fairness metrics are important for revealing cross‐domain gaps in moral foundation classification, but they capture only one axis of unfairness, for example TRP and FPR. Moreover, multi‐label tasks with skewed distribution for moral dimensions like \texttt{authority} can produce unreliable fairness estimates when labels are rare \cite{huang2021balancingmethodsmultilabeltext}. Therefore, the fairness metric for the moral sentiment classification, should detect moral foundations consistently across different social media platforms when the underlying moral content is equivalent, regardless of platform specific linguistic conventions or communication norms. In the scope of this research, the platform bias in moral classification represents a form of algorithmic unfairness for several critical reasons. But mainly it will amplify systematic discrimination or unfairness to specific platform. Therefore, this research proposes a fairness metric which focuses on the consistency for moral foundations. Moral Fairness Consistency (MFC) provides a more intuitive and explainable approach that directly measures how consistently models detect moral foundations across platforms. We define the absolute difference in fairness for each moral label \( l \in \{1, \dots, 5\} \) between cross-domain directions as:

\begin{equation}
\text{Diff}^{(l)} = \left| \text{Avg}^{(l)}_{\text{Reddit} \rightarrow \text{Twitter}} - \text{Avg}^{(l)}_{\text{Twitter} \rightarrow \text{Reddit}} \right|
\end{equation}

Then, we define the Moral Fairness Consistency (MFC) score as:

\begin{equation}
\text{MFC} = 1 - \frac{1}{5} \sum_{l=1}^{5} \text{Diff}^{(l)}
\end{equation}
For instance, the label \texttt{care} exhibits high cross-domain consistency, with MFC scores of 0.9556 for DistilBERT and 0.9573 for BERT. This implies that the absolute difference in detection rates between MFRC $\to$ MFTC and MFTC $\to$ MFRC is approximately 0.044 and 0.043, respectively. In contrast, the \texttt{authority} label shows the lowest MFC scores across both models, where 0.7781 for DistilBERT and 0.7727 for BERT indicating a considerably larger divergence for domain transfers. These values suggest that moral foundation labels such as \texttt{loyalty} and \texttt{care} are learned and transferred more consistently across domains, whereas \texttt{authority} suffers from significant domain specific instability. Therefore, the MFC metric captures the extent to which models preserve fairness in moral label prediction when exposed to different social contexts, offering a more interpretable and granular measure than traditional fairness metrics alone.
\begin{table}[htbp]
  \caption{Moral Fairness Consistency (MFC) per-label}
  \label{tab:mfc_per_label}
  \centering
  \begin{tabular}{llc}
    \toprule
    \textbf{Model} & \textbf{Label} & \textbf{MFC (95\% CI)} \\
    \midrule
    DistilBERT & authority  & 0.7781 (0.7741–0.7822) \\
               & care       & 0.9556 (0.9537–0.9576) \\
               & fairness   & 0.9499 (0.9472–0.9524) \\
               & loyalty    & 0.9666 (0.9647–0.9684) \\
               & non-moral  & 0.9205 (0.9179–0.9234) \\
    \midrule
    BERT       & authority  & 0.7727 (0.7681–0.7771) \\
               & care       & 0.9573 (0.9552–0.9593) \\
               & fairness   & 0.9451 (0.9429–0.9478) \\
               & loyalty    & 0.9617 (0.9599–0.9637) \\
               & non-moral  & 0.9076 (0.9047–0.9102) \\
    \bottomrule
  \end{tabular}
\end{table}

\begin{table}[htbp]
  \caption{Spearman Correlation between MFC and Other Metrics}
  \label{tab:spearman}
  \begin{tabular}{llcc}
    \toprule
    \textbf{Model} & \textbf{Metric} & $\rho$ & $p$-value \\
    \midrule
    \multirow{5}{*}{DistilBERT}
        & F1 Score      & -0.0920 & 0.6998 \\
      & Precision     &  0.1226 & 0.6065 \\
      & Recall        & -0.0981 & 0.6807 \\
      & DP Difference & \textbf{-1.0000} & \textbf{0.0000} \\
      & EO Difference & \textbf{-0.9000} & \textbf{0.0000} \\
    \midrule
    \multirow{5}{*}{BERT}
      & F1 Score      & -0.1042 & 0.6619 \\
      & Precision     &  0.0429 & 0.8574 \\
      & Recall        & -0.1410 & 0.5532 \\
      & DP Difference & \textbf{-1.0000} & \textbf{0.0000} \\
      & EO Difference & \textbf{-0.9000} & \textbf{0.0000} \\
    
    \bottomrule
  \end{tabular}
\end{table}

\subsubsection*{Novel Fairness Metric Validation Results}
Table \ref{tab:mfc_per_label} presents the MFC scores for each moral label across both models. The results reveal significant variation in moral fairness consistency across different moral foundations. The \texttt{authority} label demonstrates the lowest MFC scores for both DistilBERT (0.7781, CI: 0.7741–0.7822) and BERT (0.7727, CI: 0.7681–0.7771), indicating substantial inconsistency in cross-domain predictions for this moral dimension. This aligns with our previous findings showing \texttt{authority} had the largest demographic parity and equalized odds differences. In contrast, \texttt{loyalty} achieved the highest MFC scores for both models (DistilBERT: 0.9666, CI: 0.9647–0.9684 and BERT: 0.9617, CI: 0.9599–0.9637), suggesting more consistent moral detection across platforms. The \texttt{care} and \texttt{fairness} labels showed moderate consistency with MFC scores around 0.94-0.96, while \texttt{non-moral} exhibited intermediate consistency (DistilBERT: 0.9205 and BERT: 0.9076).

Table \ref{tab:spearman} presents the Spearman correlation analysis between our proposed Moral Fairness Consistency metric and established performance and fairness measures. This analysis serves as a crucial validation step for understanding how MFC relates to existing evaluation frameworks and whether it captures distinct dimensions of model behavior in moral classification tasks. The novel MFC metric exhibits weak associations towards the traditional performance metrics. For DistilBERT, MFC correlations with performance metrics were: F1 Score ($\rho$ = -0.092, $\rho$ = 0.700), Precision ($\rho$ = 0.123, $\rho$ = 0.607), and Recall ($\rho$ = -0.098, $\rho$ = 0.681). For BERT: F1 Score ($\rho$ = -0.104, $\rho$ = 0.662), Precision ($\rho$ = 0.043, $\rho$ = 0.857), and Recall ($\rho$ = -0.141, $\rho$ = 0.553). All performance metric correlations were non-significant (p > 0.05). For the fairness metrics, both models exhibited identical correlation patterns with fairness metrics. MFC showed perfect negative correlation with Demographic Parity Difference ($\rho$ = -1.000, p < 0.001) and strong negative correlation with Equalized Odds Difference ($\rho$ = -0.900, p < 0.001) for both DistilBERT and BERT.

\section{Discussion}
\label{sec:Discussion}

In this section, the findings are discussed about the existing literature, the results are interpreted, and the limitations of the study are examined. Finally, the contribution of this research to the fields of moral sentiment classification and fairness evaluation is outlined.

\subsection{Comparison with the Existing Studies}
The findings of this paper align well and extend the previous work in some key aspects. Performance degradation for the cross-domain experimentation scenarios have been consistent with the work of Guo et al. \cite{guo2023datafusionframeworkmultidomain}, where it highlighted the challenges of using heterogeneous data for learning morality paradigms. The results for the in-domain MFTC fine-tuned models outperform the MFRC models in performance metrics such as micro-F1, which aligns with the work of Hoover et al. \cite{twitter}, where it was demonstrated that Twitter's text structure and character limitations can lead to much more focused and denser morality expressions, thus models can learn better for in-domain scenarios. On the other hand, asymmetric patterns have been observed for the cross-domain scenarios, where the Twitter $\to$ Reddit models show sharp drop in recall and F1 score, while the precision remained more consistent. This pattern suggests that Twitter fine-tuned models learn highly specific moral cues and expressions which perform worse when put in cross-domain environment, thus failing to generalize in Reddits diverse and longer context patterns, showing similarity with the work of Trager et al. \cite{trager2022moral}, proving contextual richness of the Reddit platform where discussion can be much more dependent on the context of itself. As shown in Figure \ref{fig:reddit_word} and Figure \ref{fig:twitter_word} (\ref{sec:apx:first_appendix}) the Reddit posts are significantly longer and more variable in form, while Reddit’s moral labels span three distinct thematic clusters (US Politics, Everyday Morality, and French Politics) suggesting Reddit's more diverse moral discourse. This diversity may explain the asymmetric transfer patterns observed in MFRC $\to$ MFTC transfers showed minimal performance degradation (0.687 to 0.672 micro-F1 for DistilBERT), while MFTC $\to$ MFRC transfers exhibited substantial drops (0.772 to 0.623). The broader contextual training in Reddit discussions may have provided more generalizable moral representations. This finding indicates that moral sentiment classification should require significant architectural and training modifications to be able to handle cross-domain integration.
The per-label fairness analysis had revealed biases that previous studies did not cover. While MoralBERT \cite{10.1145/3677525.3678694} reports improved performance through domain adversarial training, focus was mainly on the overall accuracy metrics. Therefore, the findings of this paper highlights that for moral foundations such as \texttt{authority}, the Demographic Parity Differences of 0.22 (0.22-0.23) and Equalised Odds Difference of 0.40 (0.39-0.41) indicate fairness violations that standard accuracy metrics fail to highlight. The computed 95\% CI per-label confirms the statistical significance by proving it is not a random variation. This per-label fairness analysis in the context of the moral sentiment classification has added information where previous researches did not focus.
\subsubsection*{DistilBERT vs. BERT}
An important finding that deserves a special attention is the similar performance between the DistilBERT and the full BERT model across the experiments. The results from Table \ref{tab:overall_eval} demonstrate that DistilBERT achieves performance which is identical to BERT across both in-domain and cross-domain scenarios. For in-domain settings, DistilBERT achieved micro-F1 scores of 0.772 (MFTC) and 0.687 (MFRC), compared to BERT's 0.768 and 0.685 respectively. This minimal difference (0.004 and 0.002) falls well within statistical noise, confirming that the distillation process preserved the moral reasoning capabilities of the original model. DistilBERT in our experiments retained 100.41\% of BERT’s micro-F1 performance in-domain and 99.85\% in cross-domain settings, thus aligning with \cite{sanh2020distilbertdistilledversionbert} and even surpassing the original claim of retaining 95\% of BERTs performance. The MFC scores further confirm the pattern. For the most problematic \texttt{authority} label, DistilBERT achieved an MFC score of 0.7781 compared to BERT's 0.7727, which is a negligible difference. Similarly, for the most consistent \texttt{loyalty} label, the difference was only 0.0049 (DistilBERT: 0.9666 and BERT: 0.9617). This consistency across fairness metrics indicates that model compression through distillation does not introduce additional bias or fairness violations. 

The current results extend this validation to the specific domain of moral sentiment classification and fairness evaluation. For researchers concerned with both fairness and computational efficiency, DistilBERT represents an optimal choice that maintains moral reasoning capabilities while reducing computational costs. Although the DistilBERT performs comparably to the full BERT model, both in terms of performance and fairness, the distillation process may introduce subtle violations in fairness that the novel MFC may not capture. For example, the smaller architecture with fewer attention heads might make it harder to capture subtle morality expressions like \texttt{authority}. While the MFC metric highlights if the model is consistent across domain, but it does not explain why this shift in fairness happens. Thus, future research should explore interpretable attention visualization techniques like BERTViz \cite{vig2019multiscalevisualizationattentiontransformer}, to understand how the model assigns weights to different morality inputs.

\subsection{Novel Fairness Metric of Moral Fairness Consistency (MFC)}
One of the key novelties of this paper is the exploration of the Moral Fairness Consistency (MFC) metric as a novel metric for assessing the transformer-based model fairness in the moral sentiment classification task in cross-domain environment. We specifically evaluated the proposed metric in cross-domain scenarios, where models trained on one dataset were tested on a distinct target domain with differing moral label data distributions, to ensure that moral foundations are detected consistently across different linguistic and cultural contexts. The construction of the Moral Fairness Consistency metric is based on the the Moral Foundation Theory \cite{graham2013moral}, where the it is universal, yet the moral reasoning is based on the linguistic and cultural differences of different social media platforms.

First of all, MFC shows weak and statistically non-significant correlations with traditional performance metrics such as F1 score, Precision and Recall (Figure \ref{fig:heatmap}). This suggests that MFC score captures dimensions of model behavior not directly reflected in traditional performance metrics, confirming prior observations that model accuracy alone is insufficient to characterize fairness properties under distributional shifts \cite{barocas2023fairness}.

Secondly, the MFC metric has a perfect negative correlation with Demographic Parity Difference and a strong negative correlation with the Equalised Odds Difference. The perfect correlation with the DP Difference indicates that, within this specific experimental set-up, both metrics react similarly to domains distributional changes. However, this does not imply redundancy between the two metrics. Demographic Parity Difference measures outcome disparities between subgroups within the same domain. In contrast, MFC was specifically designed to assess how consistent subgroup predictions remain when the model is exposed to new domains with different linguistic, cultural, or contextual norms but equivalent underlying moral content.

The per-label MFC scores showcase an order of robustness for the moral foundations in cross-domain context, which highlight the importance of the insights that we receive from the MFC metric to achieve fair transformer models. For instance, the \texttt{loyalty} label achieved the highest consistency for the both models, where in DistilBERT 0.9666 and BERT 0.9617. In contrast, the label \texttt{authority} showcase the lowest MFC score where DistilBERT 0.7781 and BERT 0.7727 (Table \ref{tab:mfc_per_label}). These results indicate significant inconsistency in cross-domain moral label classification and indicate that \texttt{'authority'} related morality expressions are more platform specific indicating strong difference in linguistic and contextual patterns of each social media platform.

\subsection{Limitations and Future Work}

While this research highlights some potential novel ideas several limitations should be addressed properly. First of all, this research is constrained to the use of two social media platform datasets (Reddit and Twitter) which are specifically labeled in accordance to the Moral Foundation Theory. These two platforms are only a subset of the social media platforms that are being used today, thus limiting the morality expressions. Additionally, the various textual preprocessing steps can be evaluated further to handle domain specific character use, such as emojis and and slangs. This research intentionally kept the preprocessing of textual input to minimal, trying to keep the natural contextual richness of the inputs.

Further research in novel fairness metric design or improvements in moral sentiment analysis should focus on variety of social media platforms. Although, this is still an ongoing work because creating such MFT datasets are laborious job which require annotator training with risks of bringing more bias due to differences in annotators morality distribution. Furthermore, this paper focuses only on five moral foundation labels to make the distributions of the moral labels similar to each other for both social media domains (as seen in Figure \ref{fig:label_sparc}). Adding more labels to the fine-tunning would require significant computing power to preserve the natural morality expression distribution. Therefore, future work should focus more on preserving the natural variance in morality expressions without trading performance with computational cost. Hence, exploring options with multi-task learning frameworks by adopting a shared encoder and separate output heads for different platforms \cite{10.1145/3663363}. In terms of the the proposed novel metric, the MFC shows strong empirical validation with perfect negative correlation with the existing fairness metrics and significant difference with performance measures should be additionally re-evaluated in different contexts, for example if it could be used as bias mitigation technique in terms of moral sentiment adjustments and as well as compared to additional performance and fairness metrics. Beyond the moral sentiment classification, the MFC metric could be adapted to other fairness related NLP tasks like hate speech detection and stance detection across various platforms. Additionally, the morality expressions can be studied even further with multi-modal settings by utilizing Vision-and-Language Transformer (ViLT) \cite{kim2021viltvisionandlanguagetransformerconvolution} architecture where textual and image inputs can be leveraged to assessed morality generalizability. Therefore, previously discussed results and limitations indicate that successful moral sentiment classification related tasks may require architectural changes and adaptions strategies that would take into account the domain variance of the inputs. At this moment scaling such approaches to newer platforms like TikTok or YouTube would significantly raise the computational demands. The future work might explore such approaches as adversarial training strategies proposed in \cite{10.1145/3677525.3678694} and using the Moral Fairness Consistency metric as post-hoc fairness evaluation tool that provides interpretable signals of fairness.

\section{Conclusion}
\label{sec:Conclusion}
This research aimed to highlight a critical gap in fairness evaluation for moral sentiment classification across social media platforms. The experiments conducted in the scope of this research have shown significant asymmetric performance drops in cross-domain evaluations. These results highlight the the existence of the platform specific biases and unfairness transferred to cross-domain scenarios. This means that traditional aggregate metrics fail to reveal hidden biases and unfairness of the models in terms of moral label classification where the model is fine-tuned for multi-label classification of moral sentiments. Therefore, after computing the aggregate model performance metrics, it was concluded that using per-label performance metrics will result in a better understanding of the model behaviors. Thus, as anticipated the per-label metrics highlight insights that aggregate metrics fail to capture. For example, the \texttt{authority} label demonstrates the most severe disparities, with Demographic Parity Differences of 0.22-0.23 and Equalized Odds Differences of 0.40-0.41, indicating systematic bias in how this moral foundation is detected across platforms. The majority label of \texttt{non-moral} shows relatively stable performance, which is understandable given its the most frequent label in both of the datasets. Therefore, the more granular approach of showcasing per-label metrics reveal that lower frequency moral foundations are affected disproportionately for cross-domain applications, even though the initial label distribution is more or less the same for the two social media platforms. 

Furthermore, building on the experiments conducted and their respective results, this paper has introduced a novel fairness metric called Moral Foundation Consistency (MFC), which provides more interpretable measure of fairness in the setting of cross-domain application for moral foundation classification. The MFC metric have shown strong negative correlation with the existing fairness measures such as Demographic Parity Difference ($\rho$ = -1.000, p < 0.001) while remaining independent from the performance metrics. This indicates that the new metric is capturing a unique aspect of the model behavior and focusing on the explainability aspects of it. This novel metric shows the order of the most consistent moral labels, for example \texttt{loyalty} achieving the highest consistency of 0.96, while \texttt{authority} shows the lowest consistency of 0.78.

An important finding worth mentioning is that the in-domain and cross-domain experiments revealed an interesting consistency between the BERT and DistilBERT performances across all the experimentation scenarios. The distilled version of the BERT model achieved almost identical performance as the full BERT model, showing a micro-F1 score of 0.772 versus 0.768 for MFTC and 0.687 versus 0.685 for MFRC. The minimal difference in performance indicate that the distilled version retained the ability to identify morality cues, while reducing the training time and computational resources significantly.

In conclusion, this paper tries to improve the theoretical foundations that are needed to ensure fair transformer model developments using the Moral Foundation Theory framework. The research has highlighted specific areas for future works to build on to advance the knowledge in creating morally robust and fair models.

\section*{Declarations}

\noindent\textbf{Data Availability:}
The datasets analysed in this study are publicly available. The Moral Foundations Twitter Corpus (MFTC) can be obtained from OSF at \href{https://osf.io/k5n7y/}{https://osf.io/k5n7y/}, and the Moral Foundations Reddit Corpus (MFRC) is available on Hugging Face at \href{https://huggingface.co/datasets/USC-MOLA-Lab/MFRC}{https://huggingface.co/datasets/USC-MOLA-Lab/MFRC}.

\noindent\textbf{Consent to Publish declaration:} not applicable.

\medskip
\noindent\textbf{Funding:} no external funding was received for this study.

\medskip
\noindent\textbf{Ethics and Consent to Participate declarations:} not applicable.

\newpage
\onecolumn

\appendix
\begin{appendices}

\section{Additional Visualisations}
\label{sec:apx:first_appendix}


\begin{figure*}[h]
\Description{Something}
\centering
\includegraphics[width=0.85\textwidth]{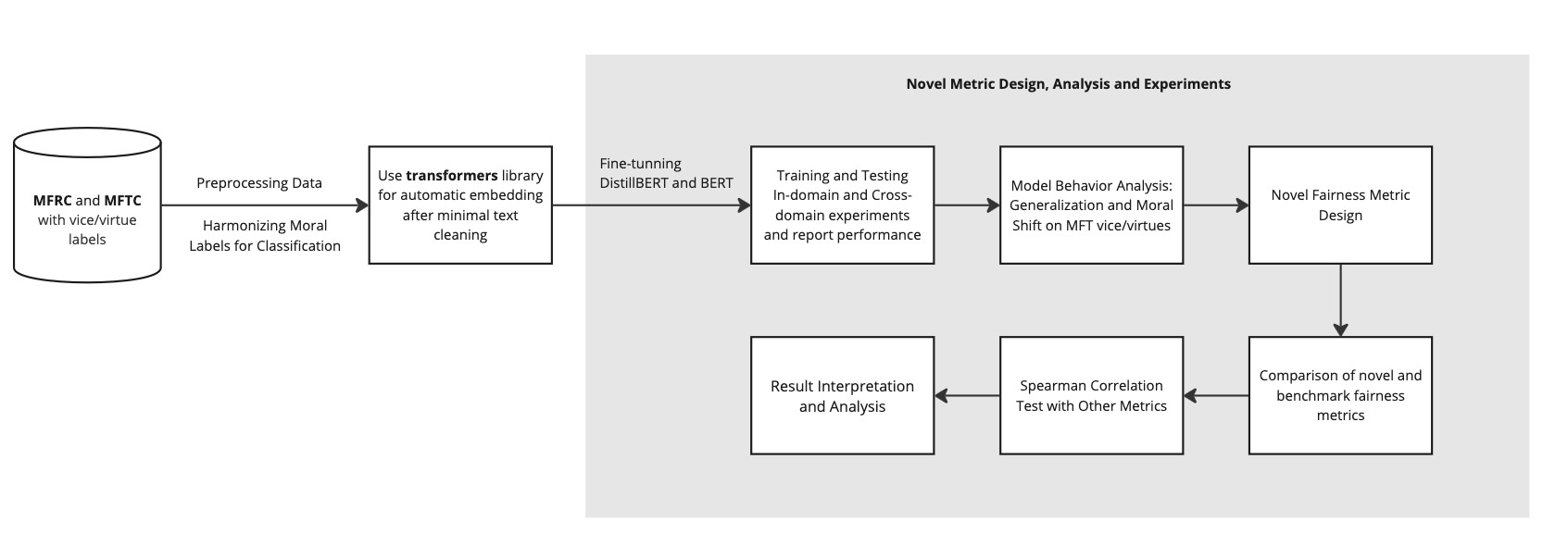}
\caption{Methodology Pipeline}
\label{fig:pipeline_new}
\end{figure*}

\begin{figure*}[h]
\Description{Something}
\centering
\includegraphics[width=0.75\textwidth]{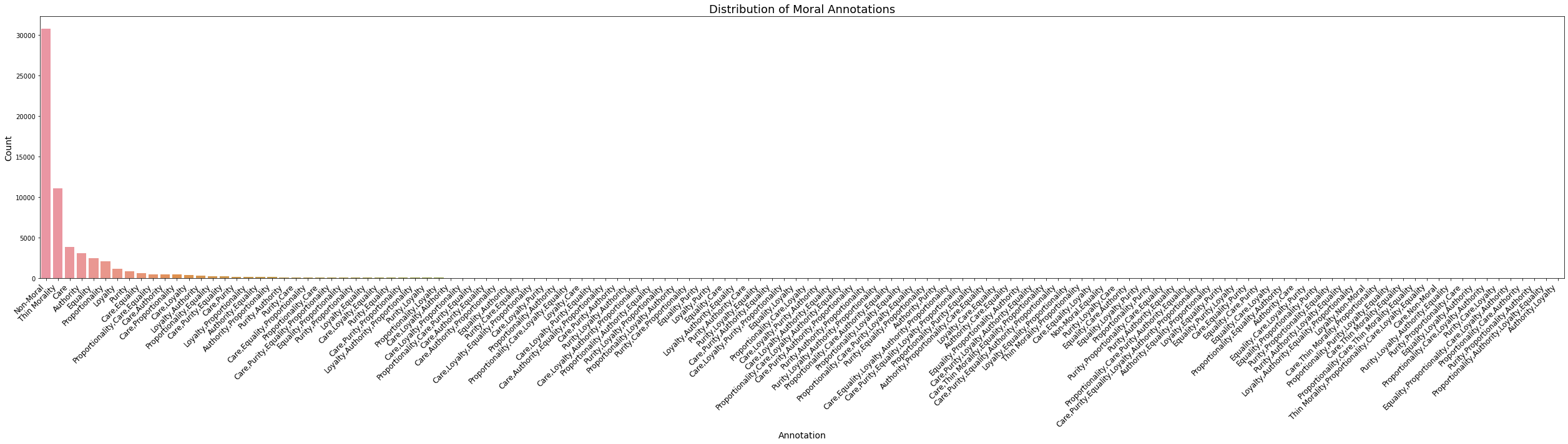}
\caption{Original Moral Label Distribution of MFRC}
\label{fig:allLABELREDDIT}
\end{figure*}

\begin{figure*}[h]
\Description{Something}
\centering
\includegraphics[width=\textwidth]{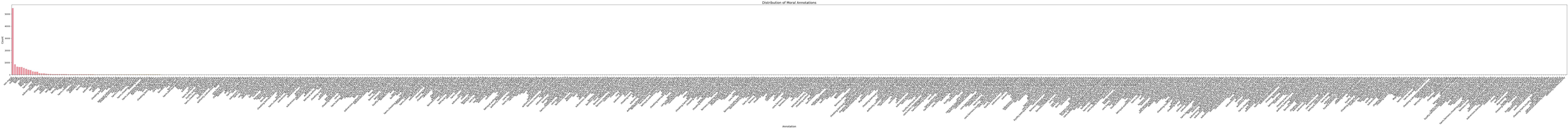}
\caption{Original Moral Label Distribution of MFTC}
\label{fig:alltwitterlables}
\end{figure*}

\begin{figure*}[h]
\Description{Something}
\centering
\includegraphics[width=0.75\textwidth]{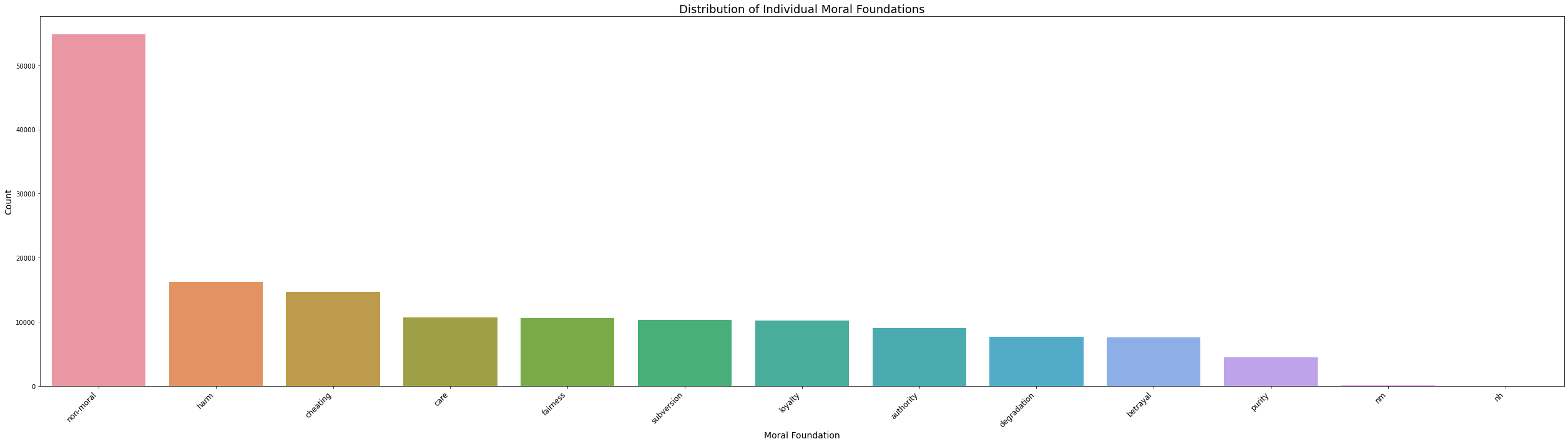}
\caption{Individual Moral Label Distribution of MFTC}
\label{fig:twiiterLABE}
\end{figure*}

\begin{figure*}[t]
\Description{Something}
\centering
\includegraphics[width=0.75\textwidth]{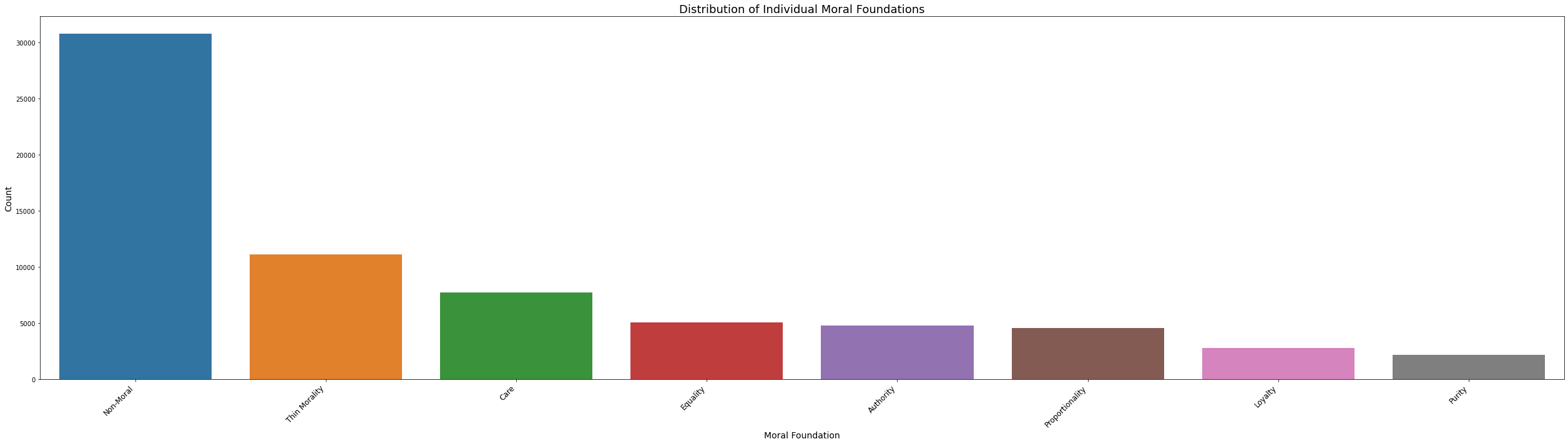}
\caption{Individual Moral Label Distribution of MFRC}
\label{fig:labelR}
\end{figure*}


\begin{figure*}[t]
\Description{Something}
\centering

\subfloat[MFTC Dataset Overview\label{fig:dataset1}]{
  \begin{minipage}{0.47\textwidth}
    \centering
    \includegraphics[width=\linewidth]{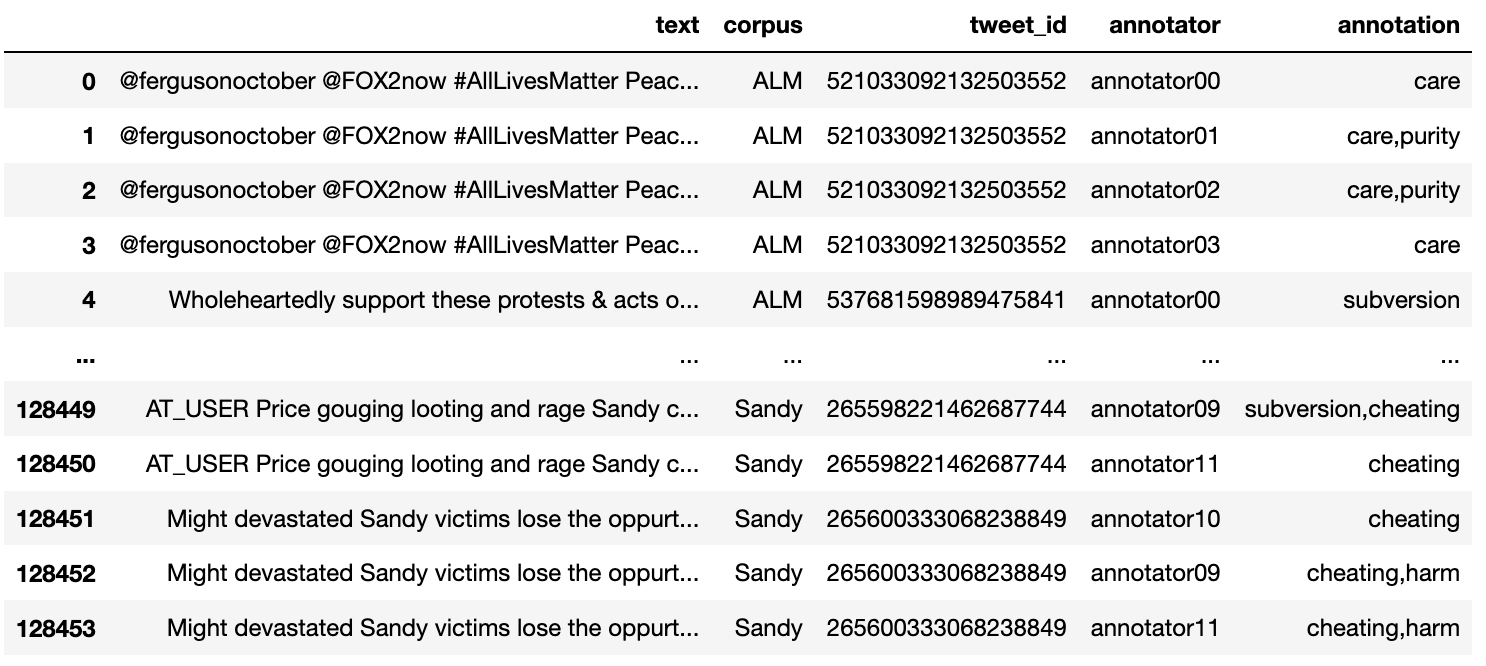}
  \end{minipage}
}
\hfill
\subfloat[MFRC Dataset Overview\label{fig:dataset2}]{
  \begin{minipage}{0.47\textwidth}
    \centering
    \includegraphics[width=\linewidth]{\detokenize{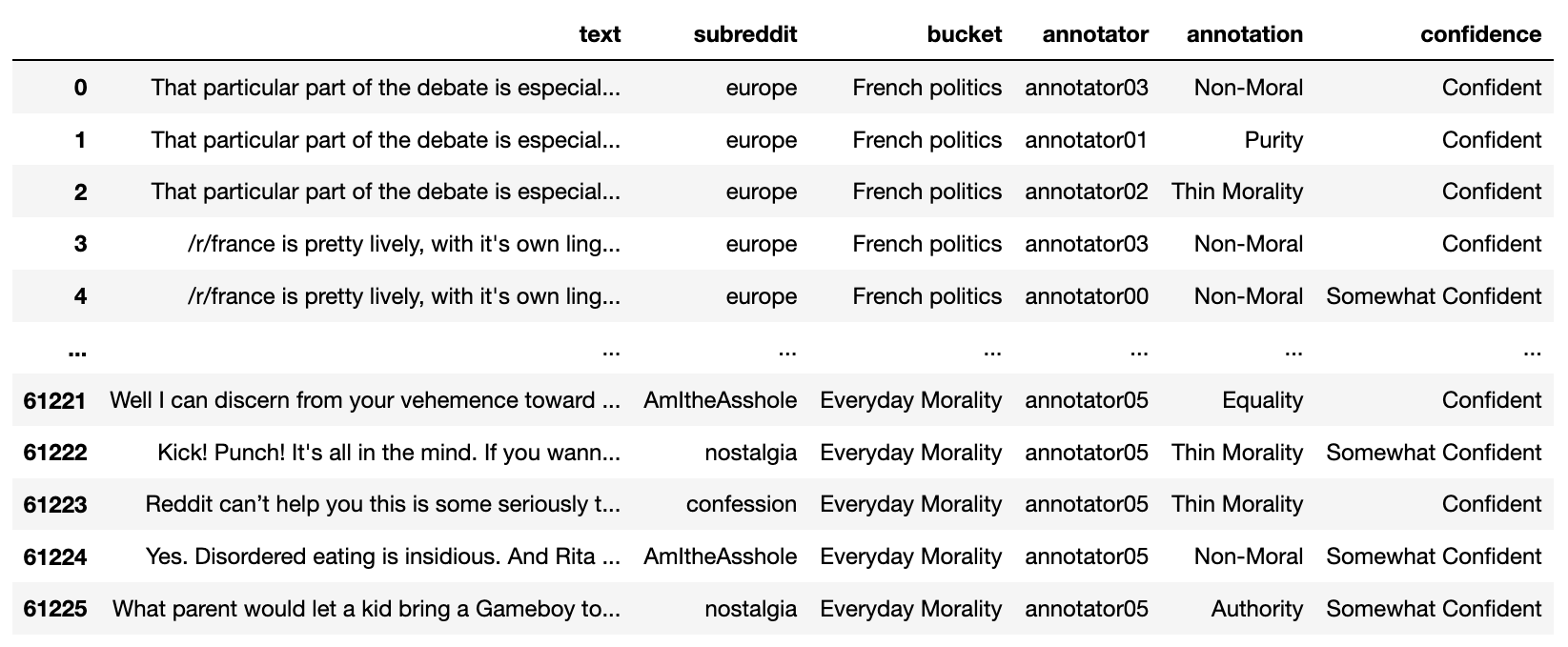}}
  \end{minipage}
}

\caption{Dataset overviews: (a) MFTC and (b) MFRC.}
\label{fig:datasets}
\end{figure*}


\begin{figure*}[t]
\Description{Something}
\centering
\includegraphics[width=0.65\textwidth]{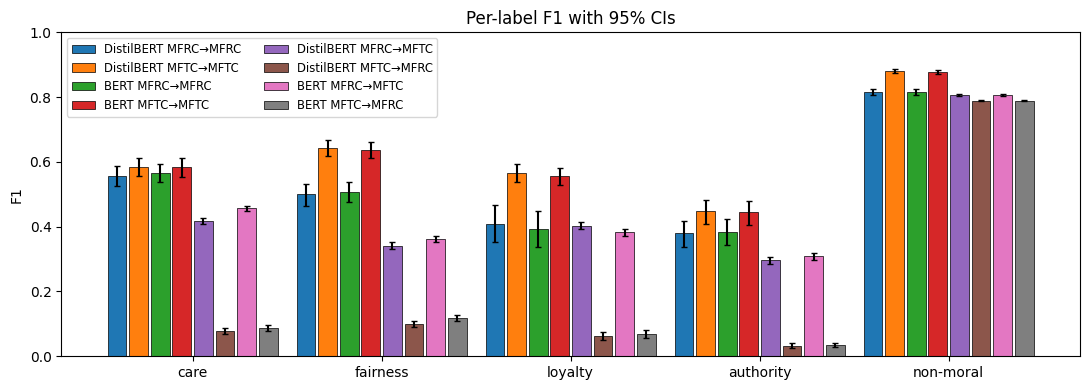}
\caption{Per-label F1 Score}
\label{fig:perlabelf1}
\end{figure*}

\begin{figure*}[t]
\Description{Something}
\centering
\includegraphics[width=0.65\textwidth]{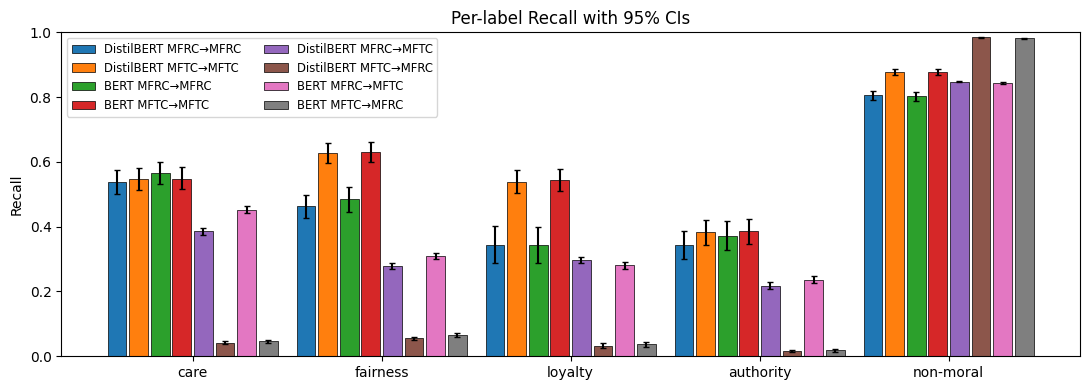}
\caption{Per-label Recall}
\label{fig:perlabelrecall}
\end{figure*}

\begin{figure*}[t]
\Description{Something}
\centering
\includegraphics[width=0.65\textwidth]{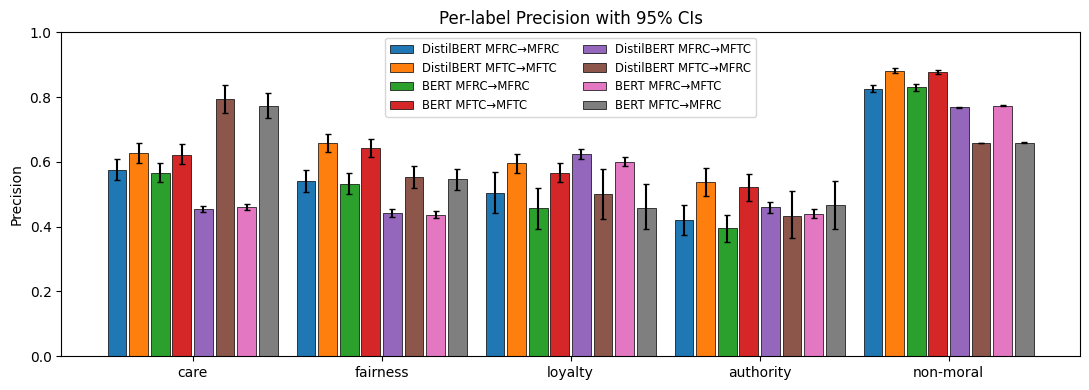}
\caption{Per-label Precision}
\label{fig:perlabelprecision}
\end{figure*}

\begin{figure*}[t]
\Description{Something}
\centering
\includegraphics[width=0.65\textwidth]{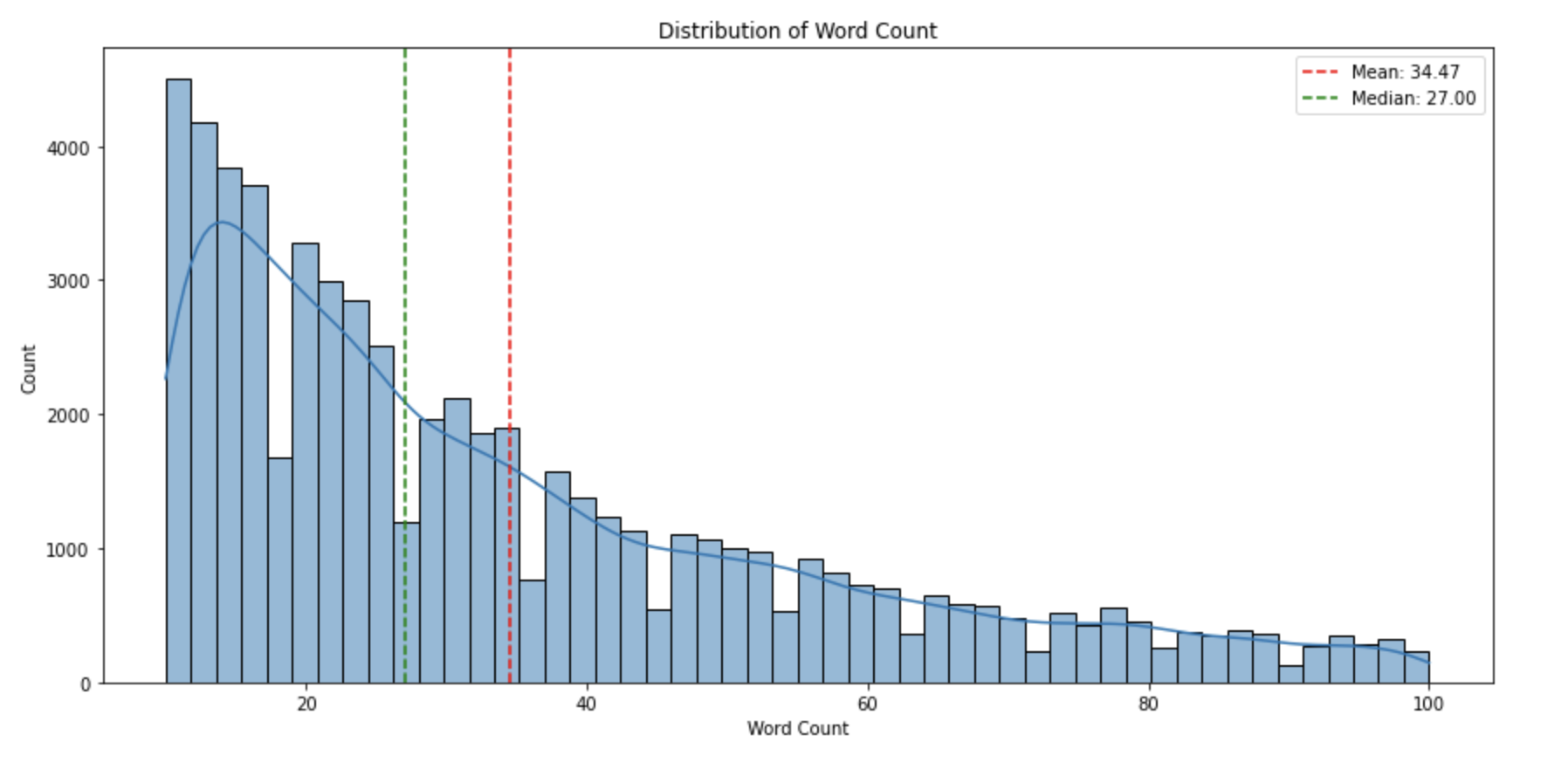}
\caption{Overall Word Count Distribution of MFRC}
\label{fig:reddit_word}
\end{figure*}

\begin{figure*}[t]
\Description{Something}
\centering
\includegraphics[width=0.65\textwidth]{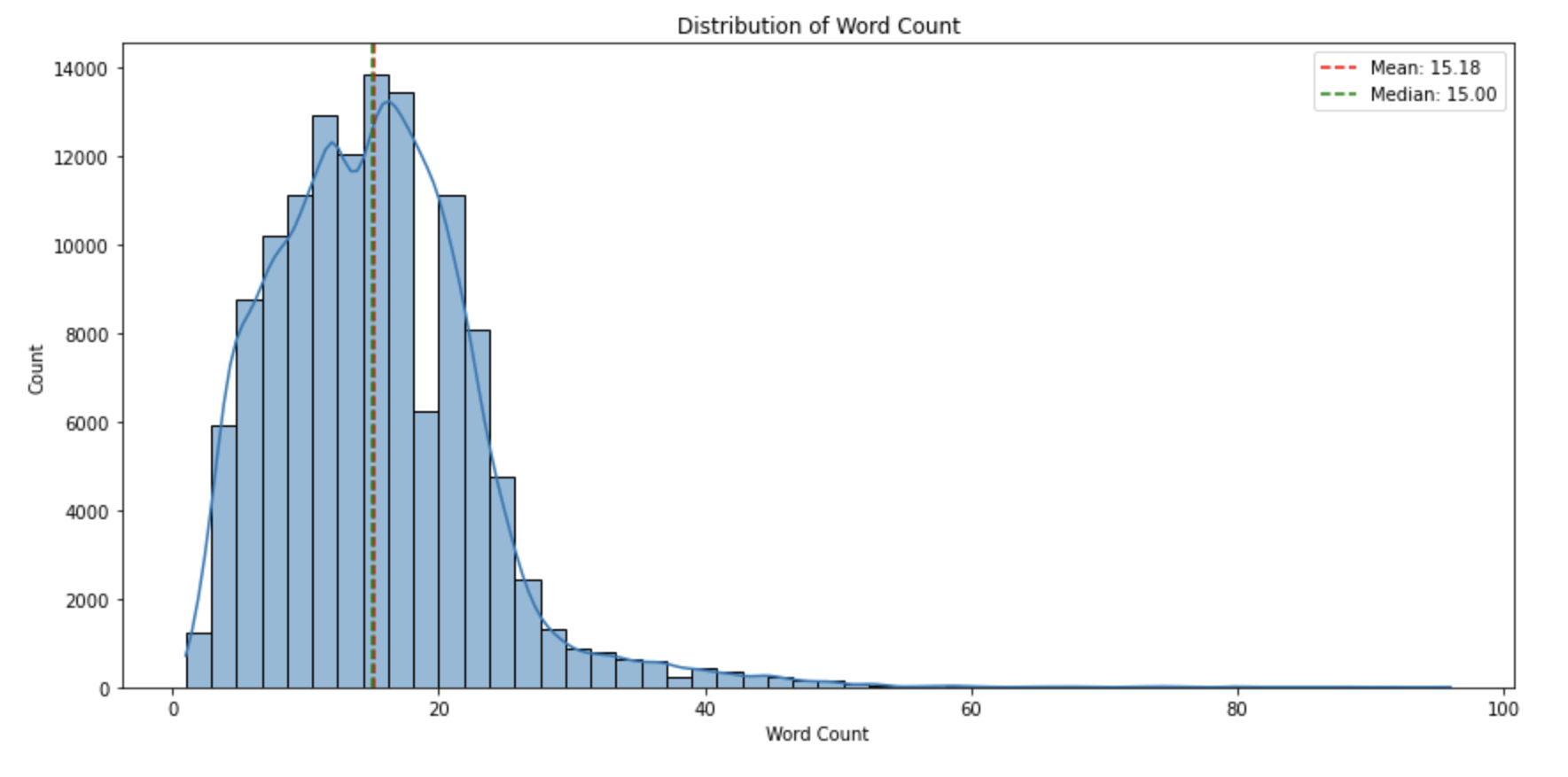}
\caption{Overall Word Count Distribution of MFTC}
\label{fig:twitter_word}
\end{figure*}

\begin{figure*}[t]
\Description{Something}
\centering
\includegraphics[width=0.65\textwidth]{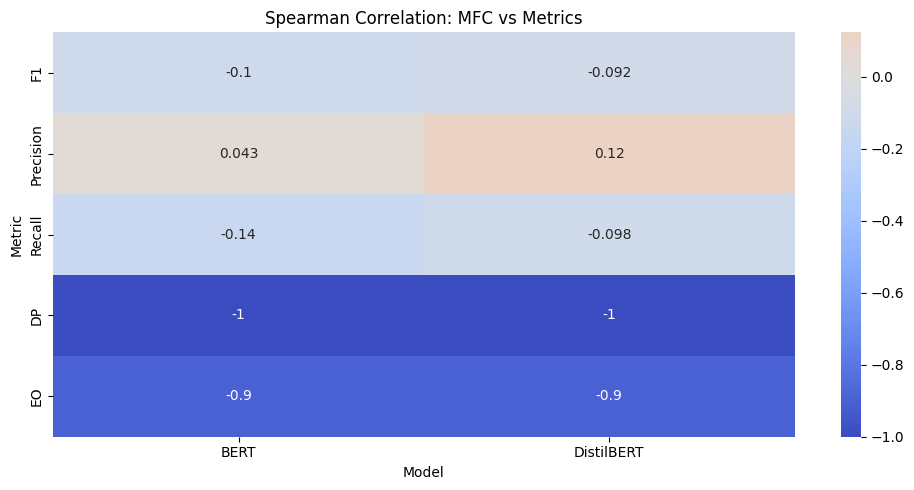}
\caption{Heatmap of Spearman correlations between MFC and selected metrics}
\label{fig:heatmap}
\end{figure*}

\begin{figure*}[t]
\Description{Something}
\centering
\includegraphics[width=0.65\textwidth]{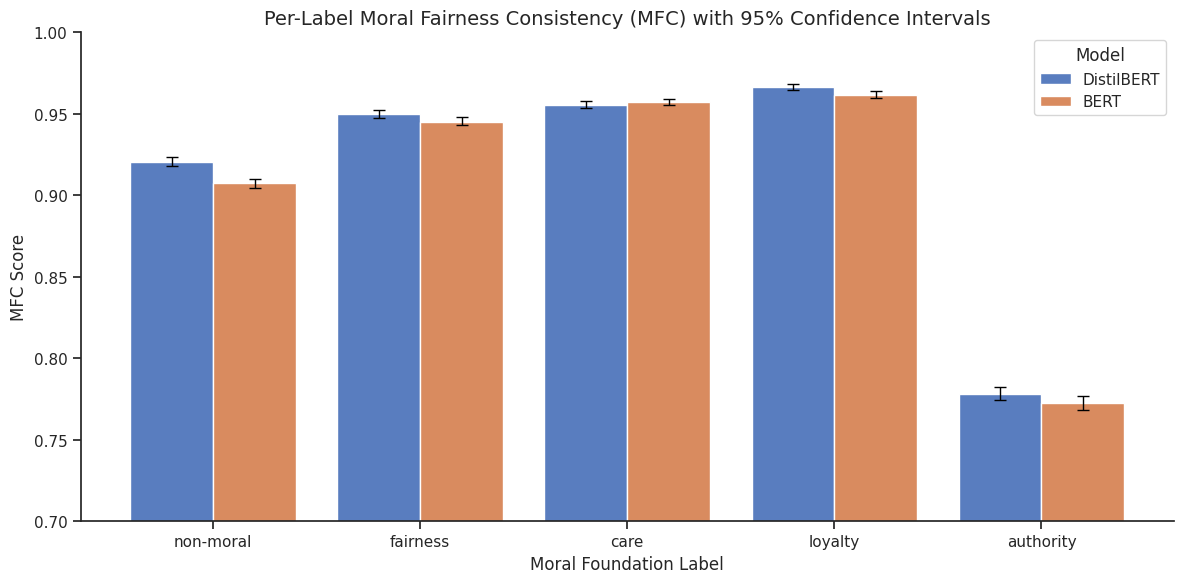}
\caption{Computed Moral Foundation Consistency with 95\% CI}
\label{fig:MFC_with_CI}
\end{figure*}

\begin{figure*}[t]
\Description{Something}
\centering
\includegraphics[width=0.65\textwidth]{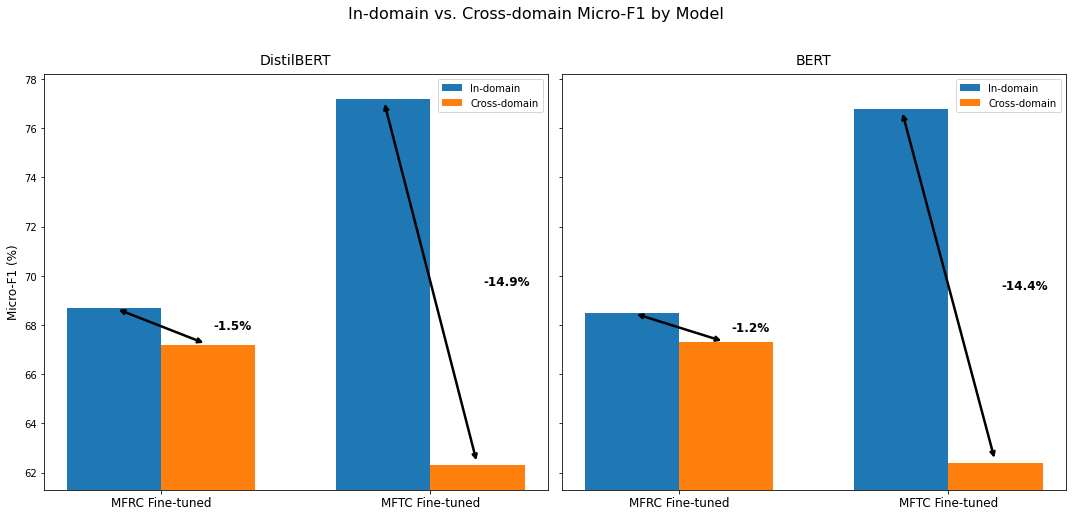}
\caption{Performance Degradation in micro-F1}
\label{fig:perf_deg}
\end{figure*}

\begin{figure*}[t]
\Description{Something}
\centering
\includegraphics[width=0.95\textwidth]{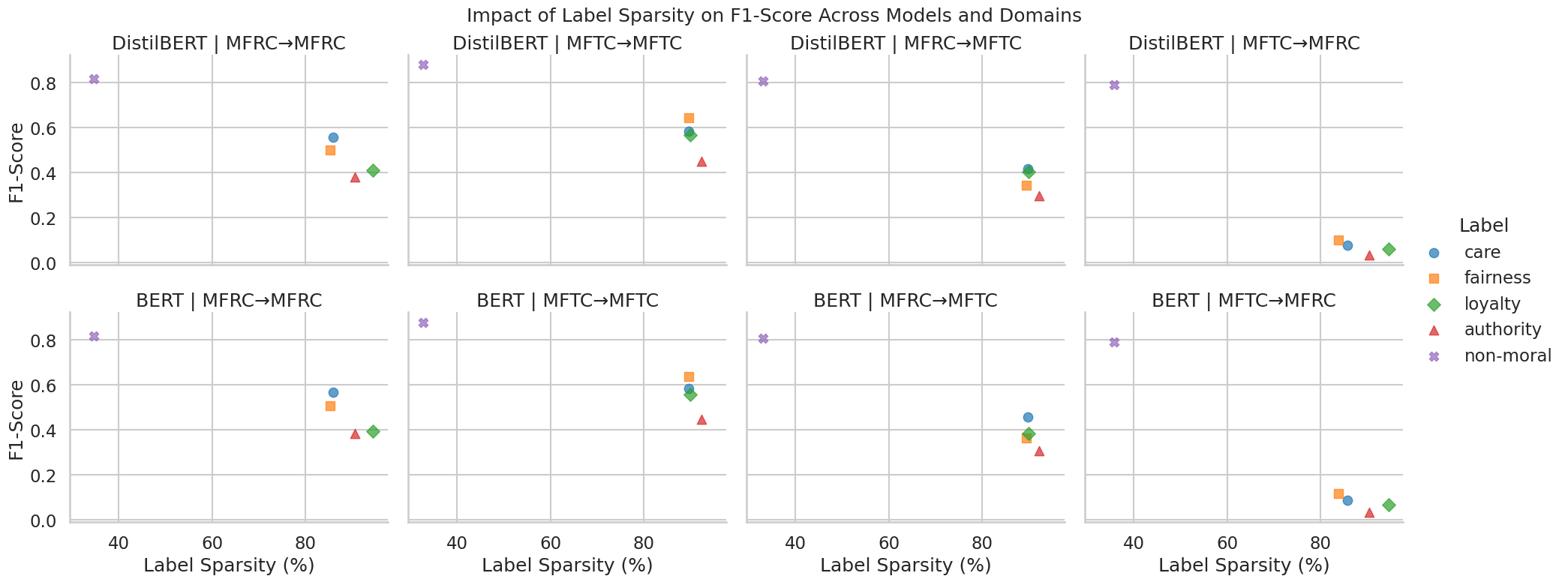}
\caption{Analysis of Label Sparsity and F1-Score}
\label{fig:label_sparc}
\end{figure*}

\end{appendices}


\begin{thebibliography}{00}

\bibitem{naveed2024comprehensiveoverviewlargelanguage}
H. Naveed, A. U. Khan, S. Qiu, M. Saqib, S. Anwar, M. Usman, N. Akhtar, N. Barnes, and A. Mian, ``A Comprehensive Overview of Large Language Models,'' arXiv:2307.06435, 2024.

\bibitem{graham2013moral}
J. Graham, J. Haidt, S. Koleva, M. Motyl, R. Iyer, S. P. Wojcik, and P. H. Ditto, ``Moral foundations theory: The pragmatic validity of moral pluralism,'' in \textit{Advances in Experimental Social Psychology}, vol. 47, pp. 55--130, 2013.

\bibitem{rogers2020primerbertologyknowbert}
A. Rogers, O. Kovaleva, and A. Rumshisky, ``A Primer in BERTology: What we know about how BERT works,'' arXiv:2002.12327, 2020.

\bibitem{ali2024understandinginterplayscaledata}
M. Ali, S. Panda, Q. Shen, M. Wick, and A. Kobren, ``Understanding the Interplay of Scale, Data, and Bias in Language Models: A Case Study with BERT,'' arXiv:2407.21058, 2024.

\bibitem{guo2023datafusionframeworkmultidomain}
S. Guo, N. Mokhberian, and K. Lerman, ``A Data Fusion Framework for Multi-Domain Morality Learning,'' arXiv:2304.02144, 2023.

\bibitem{schramowski2022largepretrainedlanguagemodels}
P. Schramowski, C. Turan, N. Andersen, C. A. Rothkopf, and K. Kersting, ``Large Pre-trained Language Models Contain Human-like Biases of What is Right and Wrong to Do,'' arXiv:2103.11790, 2022.

\bibitem{ganguli2023capacitymoralselfcorrectionlarge}
D. Ganguli \textit{et al.}, ``The Capacity for Moral Self-Correction in Large Language Models,'' arXiv:2302.07459, 2023.

\bibitem{jiang2022machineslearnmoralitydelphi}
L. Jiang \textit{et al.}, ``Can Machines Learn Morality? The Delphi Experiment,'' arXiv:2110.07574, 2022.

\bibitem{bansal2022surveybiasfairnessnatural}
R. Bansal, ``A Survey on Bias and Fairness in Natural Language Processing,'' arXiv:2204.09591, 2022.

\bibitem{Chen_2024}
S. Chen, Y. Li, S. Lu, H. Van, H. J. W. L. Aerts, G. K. Savova, and D. S. Bitterman, ``Evaluating the ChatGPT family of models for biomedical reasoning and classification,'' \textit{Journal of the American Medical Informatics Association}, vol. 31, no. 4, pp. 940--948, 2024, doi: 10.1093/jamia/ocad256.

\bibitem{openai2024gpt4technicalreport}
OpenAI \textit{et al.}, ``GPT-4 Technical Report,'' arXiv:2303.08774, 2024.

\bibitem{grattafiori2024llama3herdmodels}
A. Grattafiori \textit{et al.}, ``The Llama 3 Herd of Models,'' arXiv:2407.21783, 2024.

\bibitem{deepseekai2024deepseekv3technicalreport}
DeepSeek-AI \textit{et al.}, ``DeepSeek-V3 Technical Report,'' arXiv:2412.19437, 2024.

\bibitem{liu2019robertarobustlyoptimizedbert}
Y. Liu, M. Ott, N. Goyal, J. Du, M. Joshi, D. Chen, O. Levy, M. Lewis, L. Zettlemoyer, and V. Stoyanov, ``RoBERTa: A Robustly Optimized BERT Pretraining Approach,'' arXiv:1907.11692, 2019.

\bibitem{vaswani2017attention}
A. Vaswani, ``Attention is all you need,'' \textit{Advances in Neural Information Processing Systems}, 2017.

\bibitem{trager2022moral}
J. Trager \textit{et al.}, ``The moral foundations reddit corpus,'' arXiv:2208.05545, 2022.

\bibitem{twitter}
J. Hoover \textit{et al.}, ``Moral Foundations Twitter Corpus: A collection of 35k tweets annotated for moral sentiment,'' 2019, doi: 10.31234/osf.io/w4f72.

\bibitem{sultana2022self}
M. Sultana, M. Naseer, M. H. Khan, S. Khan, and F. S. Khan, ``Self-distilled vision transformer for domain generalization,'' in \textit{Proceedings of the Asian Conference on Computer Vision (ACCV)}, pp. 3068--3085, 2022.

\bibitem{article_machine_beh}
I. Rahwan \textit{et al.}, ``Machine behaviour,'' \textit{Nature}, vol. 568, pp. 477--486, 2019, doi: 10.1038/s41586-019-1138-y.

\bibitem{zangari2024survey}
L. Zangari, C. M. Greco, D. Picca, and A. Tagarelli, ``A Survey on Moral Foundation Theory and Pre-Trained Language Models: Current Advances and Challenges,'' arXiv:2409.13521, 2024.

\bibitem{jin2022makeexceptionsexploringlanguage}
Z. Jin, S. Levine, F. Gonzalez, O. Kamal, M. Sap, M. Sachan, R. Mihalcea, J. Tenenbaum, and B. Sch{\"o}lkopf, ``When to Make Exceptions: Exploring Language Models as Accounts of Human Moral Judgment,'' arXiv:2210.01478, 2022.

\bibitem{sanh2020distilbertdistilledversionbert}
V. Sanh, L. Debut, J. Chaumond, and T. Wolf, ``DistilBERT, a distilled version of BERT: smaller, faster, cheaper and lighter,'' arXiv:1910.01108, 2020.

\bibitem{10.1162/tacl_a_00425}
P. Czarnowska, Y. Vyas, and K. Shah, ``Quantifying Social Biases in NLP: A Generalization and Empirical Comparison of Extrinsic Fairness Metrics,'' \textit{Transactions of the Association for Computational Linguistics}, vol. 9, pp. 1249--1267, 2021, doi: 10.1162/tacl\_a\_00425.

\bibitem{devlin2019bertpretrainingdeepbidirectional}
J. Devlin, M.-W. Chang, K. Lee, and K. Toutanova, ``BERT: Pre-training of Deep Bidirectional Transformers for Language Understanding,'' arXiv:1810.04805, 2019.

\bibitem{tan-etal-2022-domain}
Q. Tan, R. He, L. Bing, and H. T. Ng, ``Domain Generalization for Text Classification with Memory-Based Supervised Contrastive Learning,'' in \textit{Proceedings of the 29th International Conference on Computational Linguistics (COLING)}, Gyeongju, Republic of Korea, pp. 6916--6926, Oct. 2022.

\bibitem{10.1145/3677525.3678694}
V. Preniqi, I. Ghinassi, J. Ive, C. Saitis, and K. Kalimeri, ``MoralBERT: A Fine-Tuned Language Model for Capturing Moral Values in Social Discussions,'' in \textit{Proceedings of the 2024 International Conference on Information Technology for Social Good (GoodIT '24)}, pp. 433--442, 2024, doi: 10.1145/3677525.3678694.

\bibitem{endalie2025fine}
D. Endalie, ``Fine-Tuning BERT Models for Multiclass Amharic News Document Categorization,'' \textit{Complexity}, vol. 2025, no. 1, p. 1884264, 2025.

\bibitem{bird2020fairlearn}
S. Bird, M. Dud{\'\i}k, R. Edgar, B. Horn, R. Lutz, V. Milan, M. Sameki, H. Wallach, and K. Walker, ``Fairlearn: A toolkit for assessing and improving fairness in AI,'' Microsoft, Tech. Rep. MSR-TR-2020-32, 2020.

\bibitem{carriero2024harmsclassimbalancecorrections}
A. Carriero, K. Luijken, A. de Hond, K. G. M. Moons, B. van Calster, and M. van Smeden, ``The harms of class imbalance corrections for machine learning based prediction models: a simulation study,'' arXiv:2404.19494, 2024.

\bibitem{wang-etal-2022-simkgc}
L. Wang, W. Zhao, Z. Wei, and J. Liu, ``SimKGC: Simple Contrastive Knowledge Graph Completion with Pre-trained Language Models,'' in \textit{Proceedings of the 60th Annual Meeting of the Association for Computational Linguistics (Long Papers)}, Dublin, Ireland, pp. 4281--4294, May 2022, doi: 10.18653/v1/2022.acl-long.295.

\bibitem{huang2021balancingmethodsmultilabeltext}
Y. Huang, B. Giledereli, A. K{\"o}ksal, A. {\"O}zg{\"u}r, and E. Ozkirimli, ``Balancing Methods for Multi-label Text Classification with Long-Tailed Class Distribution,'' arXiv:2109.04712, 2021.

\bibitem{tiwari2025advancingvulnerabilityclassificationbert}
H. Tiwari, ``Advancing Vulnerability Classification with BERT: A Multi-Objective Learning Model,'' arXiv:2503.20831, 2025.

\bibitem{barocas2023fairness}
S. Barocas, M. Hardt, and A. Narayanan, ``Fairness and machine learning: Limitations and opportunities,'' MIT Press, 2023.

\bibitem{sheskin2003handbook}
D. J. Sheskin, ``Handbook of parametric and nonparametric statistical procedures,'' Chapman and Hall/CRC, 2003.

\bibitem{zangari-etal-2025-me2}
L. Zangari, C. M. Greco, D. Picca, and A. Tagarelli, ``ME\textsuperscript{2}-BERT: Are Events and Emotions what you need for Moral Foundation Prediction?,'' in \textit{Proceedings of the 31st International Conference on Computational Linguistics (COLING)}, Abu Dhabi, UAE, pp. 9516--9532, Jan. 2025.

\bibitem{khan2025natural}
Z. Khan \textit{et al.}, ``Natural Language Processing Techniques for Automated Content Moderation,'' \textit{International Journal of Web of Multidisciplinary Studies}, vol. 2, no. 2, pp. 21--27, 2025.

\bibitem{vida2023valuesethicsmoralsuse}
K. Vida, J. Simon, and A. Lauscher, ``Values, Ethics, Morals? On the Use of Moral Concepts in NLP Research,'' arXiv:2310.13915, 2023.

\bibitem{kiesel-etal-2022-identifying}
J. Kiesel, M. Alshomary, N. Handke, X. Cai, H. Wachsmuth, and B. Stein, ``Identifying the Human Values behind Arguments,'' in \textit{Proceedings of the 60th Annual Meeting of the Association for Computational Linguistics (Long Papers)}, Dublin, Ireland, pp. 4459--4471, May 2022, doi: 10.18653/v1/2022.acl-long.306.

\bibitem{inproceedings}
R. Rzepka and K. Araki, ``Polarization of consequence expressions for an automatic ethical judgment based on moral stages theory,'' in \textit{Proceedings}, 2012.

\bibitem{radaideh2025fairness}
M. I. Radaideh, O. H. Kwon, and M. I. Radaideh, ``Fairness and social bias quantification in Large Language Models for sentiment analysis,'' \textit{Knowledge-Based Systems}, p. 113569, 2025.

\bibitem{uddin2024novel}
S. Uddin, H. Lu, A. Rahman, and J. Gao, ``A novel approach for assessing fairness in deployed machine learning algorithms,'' \textit{Scientific Reports}, vol. 14, no. 1, p. 17753, 2024.

\bibitem{park2024moralitynonbinarybuildingpluralist}
J. Park, E. Liscio, and P. K. Murukannaiah, ``Morality is Non-Binary: Building a Pluralist Moral Sentence Embedding Space using Contrastive Learning,'' arXiv:2401.17228, 2024.

\bibitem{vig2019multiscalevisualizationattentiontransformer}
J. Vig, ``A Multiscale Visualization of Attention in the Transformer Model,'' arXiv:1906.05714, 2019.

\bibitem{10947681}
S. Ramakrishnan and L. D. Dhinesh Babu, ``Improving Multi-Label Emotion Classification on Imbalanced Social Media Data With BERT and Clipped Asymmetric Loss,'' \textit{IEEE Access}, vol. 13, pp. 60589--60601, 2025, doi: 10.1109/ACCESS.2025.3557091.

\bibitem{kim2021viltvisionandlanguagetransformerconvolution}
W. Kim, B. Son, and I. Kim, ``ViLT: Vision-and-Language Transformer Without Convolution or Region Supervision,'' arXiv:2102.03334, 2021.

\bibitem{10.1145/3663363}
S. Chen, Y. Zhang, and Q. Yang, ``Multi-Task Learning in Natural Language Processing: An Overview,'' \textit{ACM Computing Surveys}, vol. 56, no. 12, art. 295, 2024, doi: 10.1145/3663363.


\end{thebibliography}
\end{document}